%% file: main.tex
\documentclass[10pt,twocolumn,letterpaper]{article}

\usepackage{cvpr}              

\input{preamble}

\definecolor{cvprblue}{rgb}{0.21,0.49,0.74}
\usepackage[pagebackref,breaklinks,colorlinks,allcolors=cvprblue]{hyperref}
\def\paperID{*****} 
\def\confName{CVPR}
\def\confYear{2025}

\title{Securing the Skies: A Comprehensive Survey on Anti-UAV Methods, Benchmarking, and Future Directions}

\author{
Yifei Dong\textsuperscript{1}\thanks{These authors contributed equally to this work.}\ \quad
Fengyi Wu\textsuperscript{1}\footnotemark[1]\ \quad
Sanjian Zhang\textsuperscript{1}\footnotemark[1]\ \quad
Guangyu Chen\textsuperscript{1}\footnotemark[1]\ \quad
Yuzhi Hu\textsuperscript{1}\footnotemark[1]\ \\
Masumi Yano\textsuperscript{1} \quad
Jingdong Sun\textsuperscript{2} \quad
Siyu Huang\textsuperscript{3} \quad
Feng Liu\textsuperscript{4} \quad
Qi Dai\textsuperscript{5} \quad
Zhi-Qi Cheng\textsuperscript{1}\footnotemark[1]\, \thanks{Corresponding author (zhiqics@uw.edu).}\\[6pt]
\textsuperscript{1}University of Washington \quad
\textsuperscript{2}Carnegie Mellon University\\
\textsuperscript{3}Clemson University \quad
\textsuperscript{4}Drexel University \quad
\textsuperscript{5}Microsoft Research\\[2pt]
{\tt\small \{yfeidong, fyiwu, sanjian, gychen, masumi76, zhiqics\}@uw.edu, hyz0929@bu.edu}\\
{\tt\small  jingdons@cs.cmu.edu, siyuh@clemson.edu, fl397@drexel.edu, qid@microsoft.com}
}

\begin{document}
\maketitle

\input{sec/0-abstract}

\input{sec/tab-1}
\input{sec/1-introduction}

\input{sec/tab-2}
\input{sec/2-datasets}
\input{sec/tab-3}
\input{sec/3-recognition}

\input{sec/tab-4}
\input{sec/4-detection}

\input{sec/fig-2}

\input{sec/5-tracking}

\input{sec/6-emerging_future}
\input{sec/9-conclusion}

{
    \small
    \bibliographystyle{ieeenat_fullname}
    \bibliography{main}
}


\end{document}

%% file: preamble.tex
\usepackage{times}
\usepackage{epsfig}
\usepackage{graphicx}
\usepackage{amsmath,amssymb}
\usepackage{array}

\newcolumntype{C}[1]{>{\centering\arraybackslash}p{#1}}

\usepackage{booktabs}      
\usepackage{tabularx}      
\usepackage{threeparttable}
\usepackage{multirow}
\usepackage{caption}
\usepackage{xspace}     
\usepackage{subcaption}  
\usepackage{makecell} 
\usepackage{adjustbox}  
\usepackage{ragged2e}    
\usepackage{array}

\usepackage{graphicx}
\usepackage{tikz}
\usepackage[edges]{forest}
\usetikzlibrary{arrows.meta,shadows.blur}

%% file: sec/0-abstract.tex
\begin{abstract}
Unmanned Aerial Vehicles (UAVs) are indispensable for infrastructure inspection, surveillance, and related tasks, yet they also introduce critical security challenges. This survey provides a wide-ranging examination of the anti-UAV domain, centering on three core objectives—classification, detection, and tracking—while detailing emerging methodologies such as diffusion-based data synthesis, multi-modal fusion, vision-language modeling, self-supervised learning, and reinforcement learning. We systematically evaluate state-of-the-art solutions across both single-modality and multi-sensor pipelines (spanning RGB, infrared, audio, radar, and RF) and discuss large-scale as well as adversarially oriented benchmarks. Our analysis reveals persistent gaps in real-time performance, stealth detection, and swarm-based scenarios, underscoring pressing needs for robust, adaptive anti-UAV systems. By highlighting open research directions, we aim to foster innovation and guide the development of next-generation defense strategies in an era marked by the extensive use of UAVs.
\end{abstract}

%% file: sec/tab-1.tex
\begin{table*}[htbp]
\centering
\caption{\small \textbf{Statistical Comparison of Public UAV Datasets.} 
   ``Seq.'' = number of sequences, 
   ``Frame'' = number of frames,
   ``T-Type'' = target types, 
   ``S-Type'' = scenario types.
   ``Modality'': RGB, IR, Audio (A), Radar (RA), LiDAR (L), RF.
   ``Task'': Classification (C), Detection (D), Tracking (T).
   ``Annotation'': class, mask, bbox, centroid, 3D coords (3D coordinates), or trajectory.}
\vspace{-0.5em}
\renewcommand\arraystretch{1.1}
\resizebox{\linewidth}{!}{
\begin{tabular}{l c c c c c c c c c c}
\toprule[1.5pt]
\textbf{Dataset} & \textbf{Task} & \textbf{Modality} & \textbf{Seq.} & \textbf{Frame} 
& \textbf{Annotation} & \textbf{Resolution} & \textbf{Rate (Hz)} & \textbf{T-Type} 
& \textbf{S-Type} & \textbf{Venue (Year)} \\
\midrule

\rowcolor{gray!20}
\textbf{FL-Drones \cite{rozantsev2016detecting}} 
& D, T & RGB
& 14 & 39K 
& Bbox
& 640$\times$480 / 752$\times$480
& --
& 1 & 2 
& TPAMI (2016) \\

\textbf{NPS-Drones \cite{li2016multi}} 
& D, T & RGB
& 50 & 70K 
& Bbox
& 1920$\times$1280 / 1280$\times$760
& --
& 1 & --
& IROS (2016) \\

\rowcolor{gray!20}
\textbf{USC Drone \cite{USC_Drone}}
& D, T & RGB
& 60 & 20K
& Bbox
& 1920$\times$1080
& 15
& 1 & --
& APSIPA ASC (2017) \\

\textbf{DroneRF \cite{allahham2019dronerf}}
& C, D & RF
& 454 & 10M
& Class
& 14-bit (RF)
& Less than 1K (RF)
& 3 & --
& Data in Brief (2019) \\

\rowcolor{gray!20}
\textbf{DroneAudio \cite{al2019droneaudio}}
& C, D & A
& --
& 1332 clips
& Class
& --
& 16K (A)
& 2 & --
& IWCMC (2019) \\

\textbf{MAV-VID \cite{mavvid2020}}
& D, T & RGB
& --
& 40K
& Bbox
& --
& --
& 1 & --
& IEEE Access (2020) \\

\rowcolor{gray!20}
\textbf{Hui \cite{Hui1}}
& D, T & IR
& 22 & 16K
& Centroid
& 256$\times$256
& 100
& 1 & 22
& CSD (2020) \\

\textbf{Acoustic-UAV \cite{2021Casabiancaacostic}}
& C, D & A
& --
& 91K sec 
& Class
& --
& 22.05K (A)
& 9 & 6
& Drones (2021) \\

\rowcolor{gray!20}
\textbf{mDrone \cite{zhao2021mdrone}}
& C, D & RA
& 100 & 60K
& 3D coords
& --
& 10 (RA)
& 1 & --
& ICRA (2021) \\

\textbf{Det-Fly \cite{zheng2021air}}
& D & RGB
& --
& 13K
& Bbox
& 3840$\times$2160
& 5
& 1 & 4
& RA-L (2021) \\

\rowcolor{gray!20}
\textbf{MOT-Fly \cite{chu2023experimental}}
& C, D & RGB
& 16 & 11K
& Bbox
& 1920$\times$1080
& --
& 3 & --
& ICUS (2021) \\

\textbf{Halmstad Drone \cite{halmstad_drone}}
& D, T & RGB, IR, A
& 650 & 203K
& Bbox
& \makecell{640$\times$512 (RGB),\\320$\times$256 (IR)}
& \makecell{60 (IR),\\30 (RGB)}
& 4 & 3
& ICPR (2021) \\

\rowcolor{gray!20}
\textbf{Drone-vs-Bird \cite{Coluccia2021DroneBird}}
& C, D & RGB
& 77 & 105K
& Bbox
& 720$\times$576 -- 3840$\times$2160
& --
& 8 & --
& AVSS (2021) \\

\textbf{Anti-UAV \cite{anti_uav}}
& D, T & RGB, IR
& 318 & 297K
& Bbox
& \makecell{1920$\times$1080 (RGB),\\640$\times$512 (IR)}
& 25
& -- & 7
& TMM (2021) \\

\rowcolor{gray!20}
\textbf{UAVSwarm \cite{wang2022uavswarm}}
& D, T & RGB
& 72 & 12K
& Bbox
& 446$\times$276 -- 1919$\times$1079
& --
& 19 & 13
& Remote Sens. (2022) \\

\textbf{Synthetic Drone \cite{Wisniewski2022dataset}}
& C & RGB
& --
& 2K
& Class
& 256$\times$256
& --
& 4 & 10
& CORD (2022) \\

\rowcolor{gray!20}
\textbf{DUT Anti-UAV \cite{dut_anti_uav}}
& D, T & RGB
& 20 & 35K
& Bbox
& 160$\times$240 -- 3744$\times$5616
& --
& 1 & 7
& T-ITS (2022) \\

\textbf{AOT \cite{aot2023}}
& D, T & IR
& 4943 & 5.9M+
& Bbox
& 2448$\times$2048
& 10
& -- & --
& AWS (2023) \\

\rowcolor{gray!20}
\textbf{Anti-UAV600 \cite{anti_uav_600}}
& D, T & IR
& 600 & 723K
& Bbox
& 640$\times$512
& 25
& 1 & --
& Preprint (2023) \\

\textbf{Anti-UAV410 \cite{Huang2023AntiUAV410}}
& D, T & IR
& 410 & 438K
& Bbox
& 640$\times$512
& 25
& 1 & 7
& TPAMI (2024) \\

\rowcolor{gray!20}
\textbf{MMAUD \cite{yuan2024mmaud}}
& C, D, T & RGB, A, RA, L
& 50 & 45K
& Trajectory
& 2560$\times$960 (RGB)
& \makecell{30 (RGB), 41.8 (A),\\15 (RA), 10 (L)}
& 6 & --
& ICRA (2024) \\

\textbf{RGBT-Tiny \cite{ying2025visible}}
& D, T & RGB, IR
& 115 & 93K
& Bbox
& 640$\times$512
& 15
& 7 & 8
& TPAMI (2025) \\

\bottomrule[1.5pt]
\end{tabular}
}
\vspace{-1mm}
\label{tab:dataset}
\end{table*}

%% file: sec/1-introduction.tex
\vspace{-0.1in}
\section{Introduction}
\label{sec:intro}


Unmanned Aerial Vehicles (UAVs) have expanding roles in civilian and military domains, offering benefits for infrastructure inspection, reconnaissance, and surveillance. Yet their rapid proliferation also introduces serious security threats, including unauthorized entry into restricted airspace and covert data gathering. As low-cost, agile drones become more prevalent, the need for robust counter-UAV (anti-UAV) technologies that can detect, track, and neutralize hostile UAVs grows increasingly urgent.

\begin{figure}
    \centering
    \includegraphics[width=0.94\linewidth]{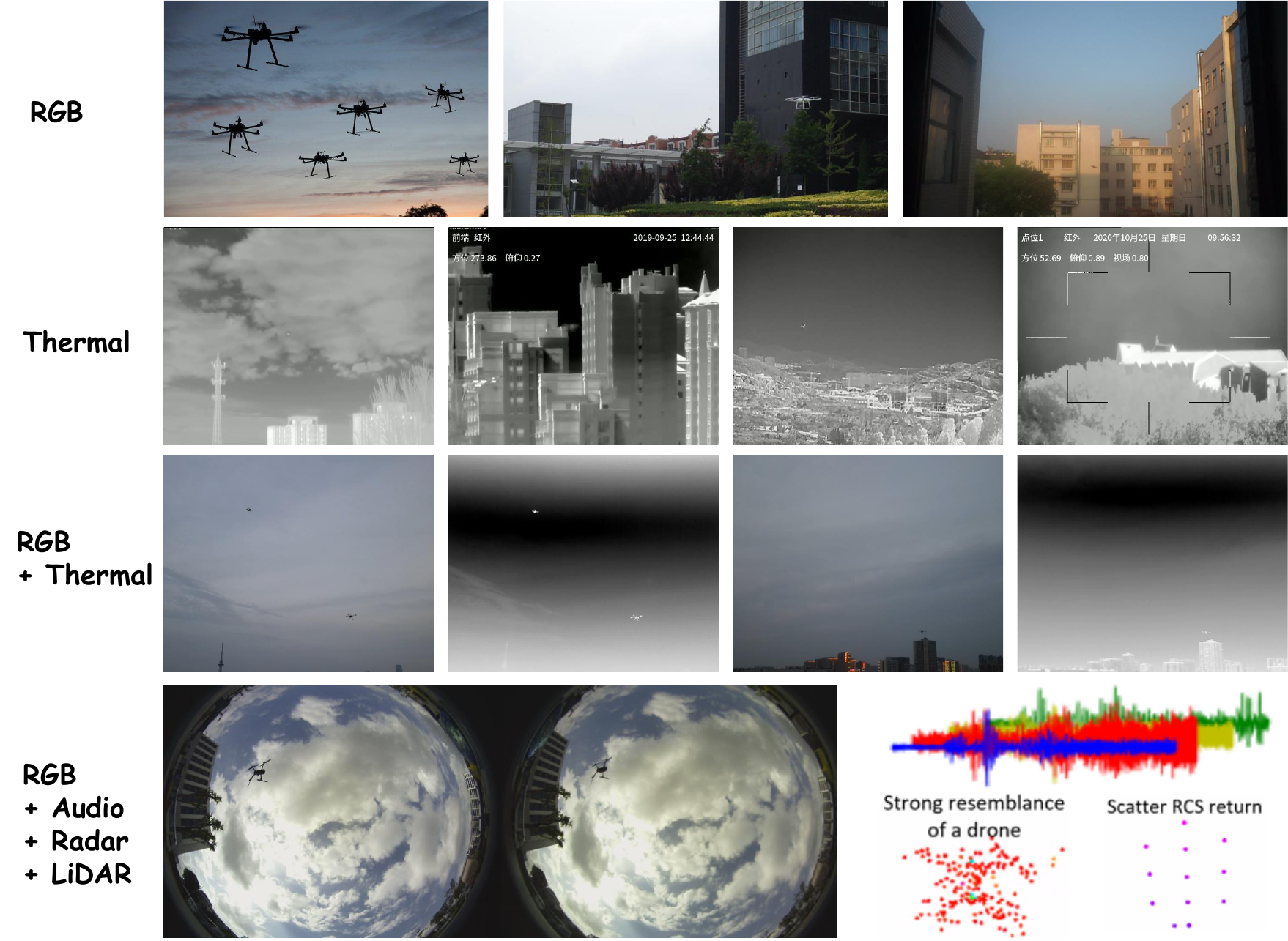}
    \vspace{-0.1in}
    \caption{\small Datasets capturing UAV with diverse modalities: 1). RGB  \cite{dut_anti_uav}, 2). Thermal \cite{Huang2023AntiUAV410}; 3). RGB + Thermal \cite{ying2025visible}; 4). RGB + Audio + Radar + LiDAR \cite{yuan2024mmaud}.}
    \label{fig:uav}
    \vspace{-0.2in}
\end{figure}

\noindent \textbf{Scope \& Motivation.}
This survey provides a comprehensive overview of current anti-UAV research, focusing on three key objectives: \emph{classification} (Section~\ref{sec:recognition_survey}), \emph{detection} (Section~\ref{sec:rep_detection}), and \emph{tracking} (Section~\ref{sec:track_survey}). We underscore the importance of leveraging multi-modal data—encompassing RGB, infrared (IR), radar, RF, and audio—to address small-object detection, erratic flight trajectories, and adversarial tactics. Figure~\ref{fig:taxonomy} illustrates five primary research thrusts that drive modern anti-UAV innovations: (\emph{i}) diffusion-based data synthesis and domain adaptation, (\emph{ii}) vision-language modeling, (\emph{iii}) self-supervised and unsupervised learning, (\emph{iv}) multi-modal fusion, and (\emph{v}) reinforcement learning.

\noindent \textbf{Datasets \& Benchmarking.}~The advancement of anti-UAV systems crucially depends on high-quality, large-scale datasets. Section~\ref{sec:uav_datasets} reviews pivotal benchmarks—including \emph{DroneRF}, \emph{Halmstad Drone}, and \emph{Anti-UAV}—detailing their annotation fidelity, sampling rates, and environmental diversity.~We also highlight the pressing need for more specialized datasets to cover scenarios such as night-time operations, swarm-based threats, and real-time interception. Table~\ref{tab:dataset} contrasts these datasets, revealing persistent gaps in annotation standards, sensor synchronization, and adversarial realism, which limit standardized evaluation and hinder robust system development.

\noindent \textbf{Classification, Detection, \& Tracking.}
While detection locates UAVs in visual or sensor data, classification provides deeper insight into UAV types and flight profiles. In Section~\ref{sec:recognition_survey}, we survey recent classification methods that integrate visual and non-visual features for improved accuracy and threat assessment. Section~\ref{sec:rep_detection} examines state-of-the-art detection pipelines ranging from traditional feature-based to CNN- and transformer-based architectures, whereas Section~\ref{sec:track_survey} discusses leading tracking algorithms (e.g., YOLO-based, Siamese) designed to manage fast-moving UAVs in cluttered or adversarial settings.

\noindent \textbf{Emerging Techniques \& Future Directions.}
Section~\ref{sec:future} spotlights key emerging paradigms—such as diffusion-driven augmentation, vision-language integration, self-supervised learning, advanced multi-modal fusion, and reinforcement learning for real-time control—and discusses their tremendous potential to reshape anti-UAV capabilities. Ultimately, these methods aim to enable adaptable, real-time systems equipped to handle evolving threats through domain adaptation, robust sensor integration, and rapid policy updates. Section~\ref{sec:future} concludes with a forward-looking perspective on next-generation anti-UAV solutions, outlining how these emerging methodologies can drive future research and industrial innovation.

%% file: sec/tab-2.tex
\begin{table*}[t]
\centering
\scriptsize
\renewcommand{\arraystretch}{1.0}
\setlength{\tabcolsep}{3pt}
\caption{\small \textbf{Representative UAV Classification Approaches by Modality.}
\scriptsize
\textit{Disclaimer:} All entries summarize our own interpretation of each paper’s methodology and reported outcomes. Since authors use distinct datasets, labels, and metrics, \textbf{no direct numerical comparison} is valid.
\textit{Abbrev.:} 
UAV = Unmanned Aerial Vehicle,\,
CNN = Convolutional Neural Network,\,
RNN = Recurrent Neural Network,\,
SVM = Support Vector Machine,\,
DNN = Deep Neural Network,\,
GAN = Generative Adversarial Network,\,
LSTM = Long Short-Term Memory,\,
CLSTM = Contextual LSTM,\,
DDI = Drone Detection and Identification,\,
PSD = Power Spectral Density,\,
MFCC = Mel-Frequency Cepstral Coefficients,\,
LFCC = Linear Frequency Cepstral Coefficients,\,
SCF = Spectral Correlation Functions,\,
TWM = Time-frequency Waterfall Map,\,
GTCC = GammaTone Cepstral Coefficients,\,
SNR = Signal-Noise Ratio,\,
APE = Average Pose Error,\,
CV = Cross Validation,\,
IR = Infrared,\,
ID = Identification,\,
Radar = RA,\,
LiDAR = L.
}
\label{tab:final_table_interpretations}
\vspace{-1em}
\resizebox{\textwidth}{!}{
\begin{tabularx}{\textwidth}{m{2.8cm} m{0.8cm} m{2.8cm} m{2.3cm} m{7.7cm}}
\toprule[1.5pt]
\textbf{Ref.\,(Yr.)}  & 
\textbf{Mod.}         &
\textbf{Method}       &
\textbf{Dataset} &
\textbf{Key Observations (Our Reading)}\\
\midrule

\multicolumn{5}{@{}l}{\textcolor{gray}{\textbf{\textit{RF-Based}}}}\\

\rowcolor{RoyalBlue!7}\textbf{Al-Sa'd et al.\,(2019) \cite{AlSaad2019}}
 & RF
 & 3 DNNs (RF fingerprint)
 & DroneRF \cite{allahham2019dronerf}
 & Introduced DroneRF (open-source, multiple flight modes). Developed three DNN classifiers for UAV presence, type, and flight mode. Achieved 99.7\% (2-class), 84.5\% (4-class), 46.8\% (10-class). Forms an early baseline on DroneRF.\\

\textbf{Al-Emadi et al.\,(2020) \cite{AlEmadi2020cnn}}
& RF
& CNN
& DroneRF \cite{allahham2019dronerf}
& {Achieved 99.8\% (2-class), 85.8\% (4-class), and 59.2\% (10-class)}. Comparable to Al-Sa'd et al.~\cite{AlSaad2019} on 2-class tasks, showing CNN’s viability for RF-based identification.\\

\rowcolor{RoyalBlue!7}\textbf{Medaiyese et al.\,(2021) \cite{Medaiyese2021}}
 & RF
 & XGBoost pipeline
 & DroneRF \cite{allahham2019dronerf} 
 & Used 227 signal segments. Reached 99.96\% (2-class), 90.73\% (4-class), 70.09\% (10-class). Overall stronger than Al-Sa'd et al.~\cite{AlSaad2019} for simpler tasks, but lags on higher-class classification vs.\ more recent solutions.\\

\textbf{Li et al.\,(2021) \cite{li2021bissiam}}
& RF
& Bispectrum Siamese Network + Contrastive Learning
& DroneRF \cite{allahham2019dronerf}
& Converts UAV signals to bispectrum for unsupervised representation. {Achieved 100\% (2-class), 98.57\% (4-class), and 92.31\% (10-class)}. Gains up to 10\% over SimCLR, SimSiam, and GAN baselines, though still below XGBoost for large class sets.\\

\rowcolor{RoyalBlue!7}\textbf{Nemer et al.\,(2021) \cite{nemer2021}}
& RF
& Ensemble learning with hierarchical classification
& DroneRF \cite{allahham2019dronerf}
& Multi-stage scheme yields~99\% overall accuracy on ten-type classification. Outperforms earlier CNN approaches (e.g., \cite{AlEmadi2020cnn}) for flight-mode tasks.\\

\textbf{Inani et al.\,(2023) \cite{Inani2023}}
& RF
& XGBoost + PSD features + 1DCNN
& DroneRF \cite{allahham2019dronerf}
& Achieved 100\% (2-class), 99.82\% (4-class), 99.51\% (10-class)—current state-of-the-art on DroneRF. Improves upon \cite{Medaiyese2021} and \cite{AlSaad2019} across all class splits, leveraging PSD feature fusion with deep learning.\\

\midrule
\multicolumn{5}{@{}l}{\textcolor{gray}{\textbf{\textit{Audio-Based}}}}\\

\rowcolor{RoyalBlue!7}\textbf{Al-Emadi et al.\,(2019) \cite{al2019droneaudio}} 
& Aud.
& CNN, RNN, CRNN
& DroneAudio \cite{al2019droneaudio}
& Released DroneAudio dataset with \(\sim 1300\) drone clips. Demonstrated CNN-based acoustic ID, achieving 96.38\% (2-class) and 92.94\% (3-class) accuracy. Provided a baseline for audio-based UAV detection.\\

\textbf{Katta~et~al.\,(2022)~\cite{Katta2022}} 
& Aud. 
& CRNN, CNN, LSTM, CLSTM, Transformer
& DroneAudio \cite{al2019droneaudio}
& Benchmarked multiple neural architectures for UAV detection and ID on DroneAudio. LSTM outperformed CRNN/CNN {with 98.93\% accuracy (2-class) and 98.60\% accuracy (3-class).}\\

\rowcolor{RoyalBlue!7}\textbf{Xiao et al.\,(2025) \cite{xiao2025tame}}
 & Aud.
 & CNN+RNN (temporal)
 & MMAUD~(audio-only) \cite{yuan2024mmaud}
 & Achieved 98.0\% classification accuracy and 0.55~APE for 3D trajectory estimation with a Mamba-based audio pipeline (4-channel short melspectrograms). Demonstrated robust performance across day/night conditions.\\

\midrule
\multicolumn{5}{@{}l}{\textcolor{gray}{\textbf{\textit{Radar-Based}}}}\\

\rowcolor{RoyalBlue!7}\textbf{Mendis et al.\,(2016) \cite{mendis2016deep}}
 & Radar
 & CNN (micro-Dopp.)
 & None public
 & Utilized deep belief networks for micro-Doppler SCF signatures, identifying UAV presence within radar beam width. One of the earliest radar-based UAV classification approaches, but lacks open data for direct replication.\\

\midrule
\multicolumn{5}{@{}l}{\textcolor{gray}{\textbf{\textit{Vision-Based}}}}\\

\rowcolor{RoyalBlue!7}\textbf{Wisniewski et al.\,(2022) \cite{Wisniewski2022}}
& RGB
& DenseNet201 CNN
& Synthetic Drone~\cite{Wisniewski2022dataset}
& Achieved 92.4\% classification accuracy (distinguishing DJI Phantom, Mavic, and Inspire) with 88.8\% precision, 88.6\% recall, 88.7\% F1-score. Demonstrated synthetic data viability for UAV type classification.\\

\midrule
\multicolumn{5}{@{}l}{\textcolor{gray}{\textbf{\textit{Multi-Modal}}}}\\

\rowcolor{RoyalBlue!7}\textbf{Deng et al.\,(2024) \cite{deng2024multi}}
 & RGB, RA, L
 & Deep fusion
 & MMAUD \cite{yuan2024mmaud} 
 & Explored 4-class UAV classification in 3D pose estimation. Fused LiDAR, radar, and fisheye camera inputs via an EfficientNet-B7–based pipeline. Achieved 81.36\% classification accuracy, improving keyframe-based multi-sensor approaches.\\

\textbf{Xiao~et~al.\,(2024)~\cite{xiao2024av}} 
& RGB, Aud. 
& AV-DTEC (self-supervised audio-visual fusion)
& MMAUD \cite{yuan2024mmaud} 
& Achieved 99.3\% classification accuracy and 0.67m APE. Parallel state-space fusion boosted performance under varied illumination. Offers an alternative to full multi-sensor pipelines by only focusing on audio-visual synergy.\\

\bottomrule[1.5pt]
\end{tabularx}
}
\vspace{-2mm}
\end{table*}

%% file: sec/2-datasets.tex
\section{UAV Datasets \& Benchmarking}
\label{sec:uav_datasets}

Table~\ref{tab:dataset} summarizes publicly available UAV datasets spanning various sensor modalities (e.g., RGB, IR, audio, radar, LiDAR, and RF), annotation schemes (e.g., bounding boxes, 3D coordinates), and sampling rates. This section highlights significant dataset initiatives, explores emerging trends, and underscores unresolved challenges.

\noindent\textbf{Foundational Efforts \& Modality Expansion.}~Initial benchmarks such as \emph{FL-Drones}~\cite{rozantsev2016detecting}, \emph{NPS-Drones}~\cite{li2016multi}, and \emph{USC Drone}~\cite{USCDroneDatasetWeb,Wang2019USCDrone} provided core RGB datasets for aerial detection and tracking. Beyond the visual domain, \emph{DroneRF}~\cite{allahham2019dronerf} captures electromagnetic signatures for RF-based UAV classification, while \emph{DroneAudio} \cite{al2019droneaudio} and \emph{Acoustic-UAV} \cite{2021Casabiancaacostic} focus on acoustic features. More recent efforts extend coverage to multi-modal data: \emph{Halmstad Drone}~\cite{svanstrom2021dataset,HalmstadDatasetRepo} fuses RGB, IR, and audio, and \emph{MMAUD}~\cite{yuan2024mmaud} integrates radar and LiDAR. Community-driven competitions (e.g., the \emph{2nd} and \emph{3rd Anti-UAV} Workshops~\cite{zhao20212nd,zhao20233rd} and the \emph{4th Anti-UAV Challenge}~\cite{AntiUAVChallenge2023}) feature adversarial or cluttered scenarios that reflect real-world operational complexity, while \emph{UAVSwarm}~\cite{wang2022uavswarm} and \emph{mDrone}~\cite{zhao2021mdrone} emphasize large-scale swarm coordination or low-visibility tracking—both highlighting the need for robust multi-sensor fusion~\cite{rosner2025multimodal}.

\noindent\textbf{Complex Scenes \& Modalities.}~Recent datasets target increasingly demanding environments and more granular annotations. For instance, \emph{Anti-UAV600}~\cite{anti_uav_600} and \emph{Anti-UAV410}~\cite{Huang2023AntiUAV410} rely on infrared or thermal imaging to capture nighttime or harsh weather conditions, while \emph{Drone-vs-Bird}~\cite{Coluccia2021DroneBird} focuses on distinguishing UAVs from visually similar wildlife. Other repositories (e.g., \emph{Midgard}~\cite{viktor2020midgard} and \emph{Rosner}~\cite{rosner2025multimodal}) explore indoor or GPS-denied settings, alongside multi-view or re-identification tasks~\cite{andle2022stanford,kalra2019dronesurf}. Simultaneously, annotations have expanded from bounding boxes to 3D coordinates and comprehensive trajectory information~\cite{yuan2024mmaud}, aligning with the growing complexity of interception and multi-UAV maneuvers.

\noindent\textbf{Trends \& Open Challenges.}~Despite notable advances in dataset scope, several gaps persist. Collections like \emph{MMAUD}~\cite{yuan2024mmaud} and \emph{Halmstad Drone}~\cite{HalmstadDatasetRepo} require precise sensor calibration and synchronization, yet standardized fusion protocols remain limited. While \emph{UAVSwarm}~\cite{wang2022uavswarm} includes up to 23 drones, testing on larger fleets (50+) remains uncommon, constraining research on scaled cooperative tracking and interception. Moreover, stealth or jamming scenarios—emerging as key topics in anti-UAV forums~\cite{zhao20212nd,zhao20233rd,AntiUAVChallenge2023}—are still sparsely represented in open-source datasets. High-resolution streams (e.g., \emph{Det-Fly}~\cite{zheng2021air} and \emph{AOT}~\cite{aot2023}) also challenge onboard systems in terms of storage and processing, prompting further research into compression and hardware acceleration~\cite{andle2022stanford,aot2023}.

%% file: sec/tab-3.tex
\begin{table*}[t]
\centering
\scriptsize
\renewcommand{\arraystretch}{1.0}
\setlength{\tabcolsep}{3pt}
\caption{\small \textbf{Representative UAV Detection Approaches by Modality.}
\scriptsize
\textit{Disclaimer:} All entries summarize our own interpretation of each paper’s methodology and reported outcomes. Since authors use distinct datasets, labels, and metrics, \textbf{no direct numerical comparison} is valid.
\textit{Abbrev.:} 
UAV = Unmanned Aerial Vehicle,\, 
CNN = Convolutional Neural Network,\, 
PSD = Power Spectral Density,\,
LSTM = Long Short-Term Memory,\, 
DNN = Deep Neural Network,\, 
RNN = Recurrent Neural Network,\, 
YOLO = You Only Look Once,\, 
FMM = Feature Modeling Module,\,
FCM = Feature Capture Module,\,
LGFFM = Local-Global Feature Focusing Module,\,
EMA = Efficient Multi-scale Attention,\,
DCNv2 = Deformable Convolution v2,\,
CSDE = Cross Stage Detail Enhance,\,
CSP-PMSA = Cross Stage Partial, Partial Multi-scale,\,
DERS = Detail Enhanced Rep Shared,\,
MCJT = Multi-Consistency Joint Tracker,\,
RMCM = Robust Motion Constraint Module,\,
FSRM = Flexible Spatial Remapping Module,\,
ATUS = Adaptive Template Update Strategy,\,
MFCC = Mel Frequency Cepstrum Coefficient,\,
RCNN = Region-based CNN,\,
FPN = Feature Pyramid Network,\,
SWIN = Shifted Window Transformer,\,
DETR = Detection Transformer,\,
OEN = Object Enhancement Net,\,
IR = Infrared,\,
LiDAR = L\,
F1 = harmonic mean of precision \& recall,\,
mAP@0.5 = mean Average Precision at 0.5 IoU,\,
mAP@0.5:0.95 = mean Average Precision at IoU in [0.5:0.95],\,
AP@0.5 = Average Precision at 0.5 IoU,\,
SOTA = state-of-the-art,\,
RA = Radar.
}
\label{tab:final_table_interpretations_dectection}
\vspace{-1em}
\resizebox{\textwidth}{!}{
\begin{tabularx}{\textwidth}{m{2.8cm} m{0.8cm} m{2.8cm} m{2.3cm} m{7.7cm}}
\toprule[1.5pt]
\textbf{Ref.\,(Yr.)}  & 
\textbf{Mod.}         &
\textbf{Method}       &
\textbf{Dataset} &
\textbf{Key Observations (Our Reading)}\\
\midrule

\multicolumn{5}{@{}l}{\textcolor{gray}{\textbf{\textit{Radar-Based}}}}\\

\rowcolor{RoyalBlue!7}\textbf{Zhao et al. (2022) \cite{zhao2021mdrone}}
& RA
& CNN + LSTM
& mDrone \cite{zhao2021mdrone}
& Exploits Doppler signals from propellers for real-time detection and 3D tracking at 10\,Hz. Achieves \(\approx\)9\,cm localization error, outperforming five classic radar baselines (e.g., 3D MUSIC) by \(\approx\)66\% in accuracy.\\

\midrule
\multicolumn{5}{@{}l}{\textcolor{gray}{\textbf{\textit{Vision-Based}}}}\\

\rowcolor{RoyalBlue!7}\textbf{Seidaliyeva et al. (2020) \cite{Seidaliyeva2020}}
& RGB
& Background subtraction + MobileNetV2
& Drone-vs-Bird \cite{Coluccia2021DroneBird}, BirdDetection \cite{yoshihashi2018}
& Combines Drone-vs-Bird with a bird dataset to enhance UAV-vs-bird discrimination. Reached 0.56 F1 at IoU=0.5 (9\,FPS). Performance is improved upon by works using more advanced YOLO variants (e.g., \cite{Dadboud2021yolov5}).\\

\textbf{Dadboud et al. (2021) \cite{Dadboud2021yolov5}}
& RGB
& YOLOv5x (PANet neck, mosaic augment)
& Det-Fly \cite{zheng2021air}, Drone-vs-Bird \cite{Coluccia2021DroneBird}
& Combined Drone-vs-Bird + Det-Fly, achieving 0.98\,mAP@0.5 with 0.96 recall. Improves upon \cite{Seidaliyeva2020} on Drone-vs-Bird, though different training splits limit exact comparison.\\

\rowcolor{RoyalBlue!7}\textbf{Sangam et al. (2023) \cite{Sangam2023TransVisDrone}}
& RGB 
& CSPDarkNet-53 + VideoSwin
& FL-Drones~\cite{rozantsev2016detecting}, NPS-Drones~\cite{li2016multi}, AOT~\cite{aot2023}
& On FL-Drones: 0.89 precision, 0.85 recall, 0.87 F1, 0.84 AP@0.5. On NPS-Drones: 0.94 precision, 0.92 recall, 0.93 F1, 0.93 AP@0.5. On AOT: 0.85 precision, 0.76 recall, 0.80 F1, 0.82 AP@0.5. Comparable to Rebbapragada et al.~\cite{Rebbapragada2024C2FDrone} under similar setups.\\

\textbf{Tu et al. (2024) \cite{tu2024}}
& RGB 
& YOLOv8 + FCM + FMM 
& DUT Anti-UAV\cite{dut_anti_uav}  
& AP@0.5=0.913, AP@0.5:0.95=0.612, below the best YOLO-based methods on this dataset (e.g., \cite{huang2024EDGSYOLOv8}) but highlights a lightweight, efficient design.\\ 

\rowcolor{RoyalBlue!7}\textbf{Cheng et al. (2024) \cite{Cheng2024LocalGlobal}}
& RGB 
& YOLOv5s + LGFFM
& DUT Anti-UAV \cite{dut_anti_uav}  
& mAP@0.5=0.943. Improves the recall rate of 2.96\% and the precision of 1.41\% compared with YOLOv5s. Slightly outperforms \cite{tu2024} on the same benchmark. Low-complexity approach with 4.4\,ms/frame inference.\\

\textbf{Bo et al. (2024) \cite{bo2024yolov7gs}}
& RGB
& YOLOv7-tiny + InceptionNeXt + SPPFCSPC-SR + “Get-and-Send”
& DUT Anti-UAV \cite{dut_anti_uav}
& mAP@0.5=0.932 at 104\,FPS. Balances fast inference and high accuracy, though slightly lower mAP@0.5 than \cite{Cheng2024LocalGlobal} and \cite{huang2024EDGSYOLOv8}.\\

\rowcolor{RoyalBlue!7}\textbf{Elsayed et al. (2024) \cite{Elsayed2024LERFNetAE}}
& RGB
& YOLOv6 + LERFNet + large-kernel LAM
& DUT Anti-UAV \cite{dut_anti_uav}
& mAP@0.5=0.936, with 0.953 precision. Surpasses \cite{bo2024yolov7gs} in accuracy, albeit at presumably higher resource use than the YOLOv7-tiny approach.\\

\textbf{Huang et al. (2024) \cite{huang2024EDGSYOLOv8}}
& RGB
& YOLOv8 + ghost conv + EMA + DCNv2
& Det-Fly \cite{zheng2021air}, DUT Anti-UAV \cite{dut_anti_uav}
& On DUT Anti-UAV, yields 0.971 mAP@0.5, highest among YOLO-based reports here (\cite{tu2024,Cheng2024LocalGlobal,bo2024yolov7gs,Elsayed2024LERFNetAE}). On Det-Fly, obtains 0.934 mAP@0.5.\\

\rowcolor{RoyalBlue!7}\textbf{Sun et al. (2024) \cite{Sun2024multiyolov8}}
& IR
& YOLOv8 w/ triple-input, BiFormer attention, small-object layer
& Hui \cite{Hui1}, Anti-UAV \cite{anti_uav}, Anti-UAV410 \cite{Huang2023AntiUAV410}
& On Hui dataset: 0.943 mAP@0.5 at 72\,FPS. Generalizes well to Anti-UAV (\(0.948\) mAP@0.5) and Anti-UAV410 (\(0.927\) mAP@0.5). Demonstrates strong IR-based detection across multiple IR benchmarks.\\

\textbf{Wisniewski et al. (2024) \cite{wisniewski2024fastrcnn}} 
& RGB 
& Faster R-CNN
& MAV-VID \cite{mavvid2020}, Anti-UAV \cite{anti_uav}, Drone-vs-Bird \cite{Coluccia2021DroneBird}
&  AP@0.5=0.970 on MAV-VID, AP@0.5=0.498 on  Drone-vs-Bird, AP@0.5=0.678 on Anti-UAV.\\

\rowcolor{RoyalBlue!7}\textbf{Zhou et al. (2024) \cite{zhou2024MssamHdc}} 
& RGB 
& YOLOv8 + multi-scale spatial attention + HDC
& Det-Fly \cite{zheng2021air}
& {Achieved 0.815 precision, 0.849 recall. Gains of +0.05 precision/+0.06 recall relative to YOLOv8 baseline}. Lags behind \cite{huang2024EDGSYOLOv8} and \cite{hao2025yolov8} in absolute mAP on Det-Fly, but emphasizes robust small-object detection.\\

\textbf{Rebbapragada et al. (2024) \cite{Rebbapragada2024C2FDrone}}
& RGB 
& SWIN + DETR + OEN
& FL-Drones~\cite{rozantsev2016detecting}, NPS-Drones~\cite{li2016multi}, AOT~\cite{aot2023}
& {Matches Sangam et al.~\cite{Sangam2023TransVisDrone} in precision/recall/AP@0.5 = 0.89/0.85/0.84 on FL-Drones and 0.94/0.92/0.93 on NPS-Drones with a transformer-based pipeline}. On AOT dataset, obtains 0.85 precision, 0.76 recall, AP@0.5=0.82 comparable to \cite{Sangam2023TransVisDrone}.\\

\rowcolor{RoyalBlue!7}\textbf{Hao et al. (2025) \cite{hao2025yolov8}}
& RGB
& YOLOv8 w/ CSDE, CSP-PMSA, OmniKernel, DERS head
& Det-Fly \cite{zheng2021air}, MOT-Fly \cite{chu2023experimental}
& On MOT-Fly, hits mAP@0.5=0.949 with only 1.13M parameters. On Det-Fly, achieves 0.964 mAP@0.5—slightly exceeding \cite{huang2024EDGSYOLOv8} at 0.934, though using different training splits.\\

\midrule
\multicolumn{5}{@{}l}{\textcolor{gray}{\textbf{\textit{Multi-Modal}}}}\\

\rowcolor{RoyalBlue!7}\textbf{Svanstr{\"o}m~et~al.\,(2021)~\cite{halmstad_drone}} 
& RGB, IR, Aud. 
& YOLOv2 (visible) + MFCC (audio) + LSTM 
& Halmstad Drone \cite{halmstad_drone} 
& {Achieved F1\(\approx\)0.61–0.88 across different distance with thermal IR sensor, F1\(\approx\)0.69–0.86 with visible camera, F1\(\approx\)0.93 with audio detector }. Fusion cut false alarms vs.\ single-sensor methods, illustrating multi-modal benefits.\\

\textbf{Wu et al. (2024) \cite{Wu2024multi}}
& RGB, RA
& Attention-based Spatiotemporal Fusion Net
& None public
& Achieved 93.0\% precision, 83.7\% recall, 88.7\% mAP in occluded scenes at 20\,FPS. Outperforms vision-only detectors by fusing radar signals, though dataset not publicly released.\\

\bottomrule[1.5pt]
\end{tabularx}
}
\vspace{-2mm}
\end{table*}

%% file: sec/3-recognition.tex
\vspace{-0.1in}
\section{UAV Classification Approaches}
\label{sec:recognition_survey}

Table~\ref{tab:final_table_interpretations} synthesizes representative UAV classification methods covering radio frequency (RF), audio, radar, vision, and LiDAR modalities. These methods range from simple drone-presence detection \cite{Nguyen2017} to more nuanced classifications of UAV types, flight modes, and orientations \cite{Scholes2022}. Although comparison is difficult due to inconsistent experimental protocols, modality-specific considerations—such as spectral noise or lighting variability—motivate distinct architectural and data-processing strategies.~Below, we highlight key insights and emerging trends.

\noindent\textbf{(1) RF-Based Classification.}~RF-focused approaches exploit the electromagnetic signatures emitted by UAVs to detect, identify, and even recognize specific flight modes. Al-Sa'd et al.~\cite{AlSaad2019} introduced the DroneRF dataset, demonstrating high accuracy (99.7\%) on a 2-class task but lower performance (46.8\%) on a more granular 10-class task. Subsequent work by Al-Emadi et al.~\cite{AlEmadi2020cnn} confirmed CNNs’ viability for DroneRF, achieving 99.8\% (2-class) and 59.2\% (10-class). Recent studies have explored advanced feature engineering or ensemble learning: for instance, Nemer et al.~\cite{nemer2021} outperformed CNN-based baselines via hierarchical ensembles, while Li et al.~\cite{li2021bissiam} incorporated bispectrum representations with contrastive learning. Inani et al.~\cite{Inani2023} achieved near-perfect 10-class accuracy (99.51\%) by fusing PSD features, a 1D-CNN, and XGBoost. Despite these advancements, real-world deployments face challenges including spectral congestion, adversarial jamming, and domain shifts~\cite{Allahham2020,Elleuch2024jamming,wang2024survey}.

\noindent\textbf{(2) Audio-Based Classification.}~Audio pipelines distinguish UAVs based on rotor noise and other acoustic signatures. Al-Emadi et al.~\cite{al2019droneaudio} introduced the DroneAudio dataset, reporting 96.38\% (2-class) and 92.94\% (3-class) accuracy using CNNs. Katta et al.~\cite{Katta2022} subsequently demonstrated the benefits of LSTM, CRNN, and Transformers, with LSTM attaining 98.93\% (2-class). Meanwhile, Xiao et al.~\cite{xiao2025tame} employed a temporal CNN+RNN to handle MMAUD’s audio modality, reaching 98.0\% accuracy alongside strong pose estimation results. Noise from wind, background chatter, or multiple UAVs remains an obstacle in practical scenarios \cite{Bernardini2017,Jeon2017}, motivating research into denoising filters and multi-microphone arrays \cite{Sun2023DeepLD}.

\noindent\textbf{(3) Radar-Based Classification.}~Radar offers micro-Doppler signatures that persist under reduced visibility or nighttime conditions. Early studies (e.g., Mendis et al.~\cite{mendis2016deep}) validated radar-based UAV classification using deep belief networks, though limited dataset availability inhibited broad adoption. CNN- and LSTM-based pipelines now dominate, focusing on micro-Doppler representations to improve robustness in cluttered environments \cite{zheng2021air,wang2024survey}. Nonetheless, radar hardware cost and calibration constraints remain practical challenges \cite{wang2024survey}.

\noindent\textbf{(4)~Vision-Based~Classification.}~Computer vision pipelines target UAV recognition via optical images, often focusing on manufacturer types (e.g., DJI Phantom, Mavic). Wisniewski et al.~\cite{Wisniewski2022} achieved 92.4\% accuracy on three drone types (Phantom, Mavic, Inspire) and illustrated synthetic imagery’s viability \cite{Wisniewski2022dataset}. Other studies address issues of small object size and motion blur, incorporating advanced detectors or domain adaptation \cite{Unlu2019,Scholes2022}. While performance in controlled scenarios is high, vision-based methods degrade under extreme illumination, adverse weather, or dense clutter \cite{cabrera2019detection,dewangan2023application}.

\noindent\textbf{(5) Multi-Modal Recognition.}~Integrating multiple sensors (e.g., RF+Audio \cite{Frid2024}, Radar+LiDAR+Vision \cite{yuan2024mmaud}, or Audio+Vision \cite{xiao2024av}) can mitigate single-sensor blind spots. Deng et al.~\cite{deng2024multi} fused fisheye RGB, radar, and LiDAR data from MMAUD, reaching 81.36\% accuracy on a 4-class task, while Xiao et al.~\cite{xiao2024av} combined audio and RGB streams to achieve 99.3\% accuracy and sub-meter pose errors. Although multi-modal setups improve robustness, they also require careful calibration, synchronization, and greater computational resources \cite{yuan2024mmaud,wang2024survey}.

\noindent\textbf{Key Insights \& Challenges.}~Across these approaches \cite{dewangan2023application,zheng2021air,wang2024survey,Unlu2019,cabrera2019detection}, each modality faces inherent constraints: 
(1) \textit{RF-based solutions} excel at flight-mode or type identification \cite{AlSaad2019,Allahham2020}, but radio congestion and potential adversarial attacks are critical hurdles \cite{Elleuch2024jamming,wang2024survey}. 
(2) \textit{Audio-based methods} achieve high accuracy in controlled settings yet suffer from significant performance drops in high-noise or windy environments \cite{Bernardini2017,Jeon2017,al2019droneaudio}. 
(3) \textit{Radar-based pipelines} bypass visual occlusions but incur higher costs and calibration demands \cite{mendis2016deep,wang2024survey}. 
(4) \textit{Vision-based approaches} generally provide fine-grained recognition but degrade with lighting or weather changes \cite{Wisniewski2022,Unlu2019}. 
(5) \textit{Multi-modal fusion} can mitigate single-sensor weaknesses and reduce false positives \cite{deng2024multi,Frid2024} yet raises complexity in real-time deployment \cite{yuan2024mmaud}. 
Additionally, adversarial robustness remains an open problem, particularly where sensors or neural networks may be fooled by spoofing or deceptive signals \cite{Medaiyese2021,Katta2022}. A continuing research direction involves efficient, low-latency architectures \cite{xiao2025tame,wang2024survey} that can handle on-board or edge-based processing for rapid UAV detection and classification.

%% file: sec/tab-4.tex
\begin{table*}[t]
\centering
\scriptsize
\renewcommand{\arraystretch}{1.0}
\setlength{\tabcolsep}{3pt}
\caption{\small \textbf{Representative UAV Tracking Approaches.}
\scriptsize
\textit{Disclaimer:} All entries summarize our own interpretation of each paper’s methodology and reported outcomes. Since authors use distinct datasets, labels, and metrics, \textbf{no direct numerical comparison} is valid.
\textit{Abbrev.:} UAV = Unmanned Aerial Vehicle,\,
CNN = Convolutional Neural Network,\,
RNN = Recurrent Neural Network,\,
SVM = Support Vector Machine,\,
DNN = Deep Neural Network,\,
LSTM = Long Short-Term Memory,\,
GMM = Gaussian Mixture Model,\,
IR = Infrared,\,
TIR = Thermal Infrared,\,
ID = Identification,\,
Radar = RA,\,
LiDAR = L,\,
SOTA = State of the Art,\,
YOLO = You Only Look Once,\,
YOLOX = You Only Look Once Extended,\,
DFSC = Dual-Flow Semantic Consistency,\,
LoFTR = Detector-Free Local Feature Matching with Transformers,\,
SiamRPN++ = Siamese Region Proposal Network++,\,
SiamSTA = Siamese Spatio-Temporal Attention tracker,\,
SiamSRT = Siamese Search Region-free Tracker,\,
SiamDT = Siamese Dual-semantic Tracker,\,
SiamFusion = Siamese Modality-Fusion Tracker,\,
DCF = Discriminative Correlation Filter,\,
CDCF = Change Detection-based Correlation Filter,\,
CFRM = Channel Feature Refinement Module,\,
BYTE = Bounding-box Embedding Tracker with Association,\,
STG-RPN = Spatio-Temporal Guided Region Proposal Network,\,
ID-RCNN = Instance Discrimination Region-based CNN,\,
FPS = Frames Per Second,\,
Acc = Accuracy,\,
F1 = F1 Score,\,
TLD = Tracking-Learning-Detection,\,
DFSC = Dual-Flow Semantic Consistency,\,
ATUS = Adaptive Template Updating Strategy,\,
EECF = Eagle-Eye Correlation Filter,\,
ECA = Efficient Channel Attention,\,
STFTrack = Spatio-Temporal Focused Tracker,\,
CAM = Channel Attention Module,\,
MCJT = Multi-Consistency Joint Tracker,\,
$\alpha$-IoU = Alpha Intersection over Union,\,
Swin = Shifted Window Transformer,\,
MSE = Mean Squared Error,\,
SPLT = Self-paced Learning Tracker.
}
\label{tab:final_table_interpretations_tracking}
\vspace{-1em}
\resizebox{\textwidth}{!}{
\begin{tabularx}{\textwidth}{m{2.8cm} m{0.8cm} m{2.8cm} m{2.3cm} m{7.7cm}}
\toprule[1.5pt]
\textbf{Ref.\,(Yr.)}  & 
\textbf{Mod.}         &
\textbf{Method}       &
\textbf{Dataset} &
\textbf{Key Observations (Our Reading)}\\
\midrule

\multicolumn{5}{@{}l}{\textcolor{gray}{\textbf{\textit{Detection-Based Trackers}}}}\\

 \rowcolor{RoyalBlue!7} \textbf{Jiang et al. (2023)} \cite{anti_uav}
 & {RGB, IR}
 & {GlobalTrack + DFSC}
 & {Anti-UAV} \cite{anti_uav}
 & {Released 318-sequence RGB-T UAV tracking benchmark. Achieved the best state accuracy score of 74.33\% on the validation set and 65.41\% on the test set} \\

\textbf{Wu et al. (2024)} \cite{Wu2024EagleEye}
 & {RGB}
 & {DCF + dual-fovea + contrast + opponent-color (EECF)}
 & {Anti-UAV \cite{anti_uav}, USC-Drone \cite{USC_Drone}}
 & {Real-time EECF tracker (86 FPS) via dual-fovea design. Achieved the state accuracy score of 66.42\% on Anti-UAV, 70.45\% on USC-Drone.} \\

\rowcolor{RoyalBlue!7}\textbf{Wang et al. (2024)} \cite{wang2024target}
 & {RGB}
 & {YOLOX + Swin-T + $\alpha$-IoU + BYTE}
 & {UAVSwarm} \cite{wang2022uavswarm}
 & {Detection-driven approach with $\alpha$-IoU + ByteTrack. Achieved 78.2\% tracking accuracy on UAVSwarm test split.} \\

\textbf{Deng et al.(2024)} \cite{deng2024multi}
 & {RGB, RA, L}
 & {YOLOv9 + EfficientNet-B7 + LSTM}
 & {MMAUD} \cite{yuan2024mmaud}
 & {Intergrated YOLOv9 + EfficientNet-B7 + LSTM/Kalman for classification and tracking. Achieved 2.21 Pose MSE Loss on 3D tracking of MMAUD and ranked 1st in UG2+ Challenge.} \\

\midrule

\multicolumn{5}{@{}l}{\textcolor{gray}{\textbf{\textit{Siamese (Contrastive) Trackers}}}}\\

\rowcolor{RoyalBlue!7} \textbf{Huang et al. (2021)} \cite{Huang2021SiamSTA}
 & {IR}
 & {SiamSTA}
 & {1st~/~2nd Anti-UAV Challenges} \cite{anti_uav} \cite{zhao20212nd}
 & {Achieved 74.46 and 67.30 scores on 1st~/~2nd Anti-UAV test-dev sets, ranking 1st. Extended SiamSTA with spatio-temporal attention + motion estimation.} \\


\textbf{Li et al. (2022)} \cite{li2022dual}
 & {IR}
 & {SuperDiMP (ResNet50)}
 & {2nd Anti-UAV Challenge \cite{zhao20212nd}}
 & {Used~SuperDiMP~with~dual-branch~IR~fusion.~Achieved +5.84\%~SR and +4.52\%~PR~over~baseline.~Focused~on~contrast~inversion~robustness.} \\

\rowcolor{RoyalBlue!7}\textbf{Chen et al. (2022)} \cite{Chen2022SiamSTA}
 & {IR}
 & {SiamSTA + CDCF}
 & {1st Anti-UAV Challenge} \cite{anti_uav}
 & {SiamSTA with CDCF and 3-stage redetection. Achieved top precision in 2nd Anti-UAV Challenge. Achieved 68.26 and 76.11 scores on Anti-UAV test and validation sets. Robust on fast/small UAVs.} \\

\textbf{Zhang et al. (2023)} \cite{zhang2023modality}
 & {RGB, IR}
 & {SiamFusion}
 & {Anti-UAV} \cite{anti_uav}
 & {Achieved overall 63.37 score on Anti-UAV with 14 FPS. Developed a dual-fusion tracker with modality + decision-level fusion. Used Swin backbone + Local-Global Converter.} \\

\rowcolor{RoyalBlue!7}\textbf{Xie et al. (2023)} \cite{xie2023stftrack}
 & {IR}
 & {FPN-based Siamese + STG-RPN + ID-RCNN (STFTrack)}
 & {Anti-UAV \cite{anti_uav}}
 & {STFTrack using STG-RPN and ID-RCNN. Achieved 91.2\% precision, 66.6\% success @ 12.4 FPS. Addressed distractors and thermal crossover.} \\

\textbf{Zhang et al. (2024)} \cite{zhang2024precision}
 & {IR}
 & {OSTrack + MCJT (RMCM + FSRM + ATUS )}
 & {1st~/~2nd~/~3rd Anti-UAV Challenges} \cite{anti_uav} \cite{zhao20212nd} \cite{zhao20233rd}
 & {Achieved 69.65\% Acc, 80.45\% Acc, 65.88\% Acc and 50.28\% Acc on the 1st Anti-UAV test set, 1st Anti-UAV test-dev, 2nd Anti-UAV test-dev, and 3rd Anti-UAV validation set.  Proposed MCJT (RMCM + FSRM + ATUS) with adaptive template and multi-consistency. Ran at  44 FPS.} \\

\rowcolor{RoyalBlue!7}\textbf{Huang et al. (2024)} \cite{Huang2023SiamSRT}
 & {IR}
 & {SiamSRT + C-C RPN + S-L RCNN + TMB + SCFD}
 & {Anti-UAV} \cite{anti_uav}
 & {Achieved 71.64 and 80.04 scores on Anti-UAV testing and validation sets. Global Siamese tracker using Swin-T + consistency modules (TMB + SCFD). Avoided template degradation; outperformed multiple SOTA trackers.} \\


\textbf{Huang et al. (2024)} \cite{Huang2023AntiUAV410}
 & {IR}
 & {SiamDT (Swin Transformer + Dual-Semantic RPN + Versatile R-CNN)}
 & {Anti-UAV410} \cite{Huang2023AntiUAV410}
 & {Introduced Anti-UAV410 dataset. Proposed SiamDT with dual semantic reasoning + background suppression. Achieved 68.19 and 71.65 scores on its testing and validation sets.} \\

\midrule

\multicolumn{5}{@{}l}{\textcolor{gray}{\textbf{\textit{Hybrid (Detection + Siamese) Trackers}}}}\\


\rowcolor{RoyalBlue!7}\textbf{Cheng et al. (2022)} \cite{Cheng2022AntiUAVTracking}
 & {IR}
 & {Multi-Hybrid~Attention~+~SiamRPN++ + YOLOv5 (SiamAD)}
 & {Anti-UAV} \cite{anti_uav}
 & {SiamAD with CSAM/CAM + hierarchical discriminator + YOLOv5. 88.4\% DP @ 37.1 FPS. Improved long-term IR tracking.} \\

\textbf{Yu et al. (2023)} \cite{Yu2023UnifiedTransformer}
 & {IR}
 & {Transformer + YOLOv5 + LoFTR + GMM + OSTrack}
 & {1st~/~2nd Anti-UAV Challenges \cite{anti_uav} \cite{zhao20212nd}}
 & {Achieved 2nd place in Anti-UAV Challenge (3rd) with 68.8 score. And 75.13\% and 80.38\% accuracy on the 1st and 2nd test dev, individually. Integrated OSTrack, YOLOv5, LoFTR, GMM for multi-stage IR tracking.} \\

\bottomrule[1.5pt]
\end{tabularx}
}
\vspace{-2mm}
\end{table*}

%% file: sec/4-detection.tex
\section{UAV Detection Approaches}
\label{sec:rep_detection}

Table~\ref{tab:final_table_interpretations_dectection} summarizes UAV detection pipelines spanning radar, vision, and multi-modal solutions. Each modality addresses distinct obstacles—such as small object size, cluttered environments, and adverse weather—offering unique advantages and trade-offs. Although direct numerical comparisons are challenging due to varying datasets and evaluation protocols, these methods collectively illustrate emerging trends in detection algorithms and system design. We discuss salient insights from the literature below.

\noindent\textbf{(1) Radar-Based Detection.}~Radar methods detect UAVs by exploiting micro-Doppler signatures, providing resilience under low-visibility or harsh weather conditions \cite{zhao2021mdrone, wang2024survey}. For instance, Zhao et al.~\cite{zhao2021mdrone} achieved decimeter-level localization at 10\,Hz by combining CNNs with LSTMs, outperforming conventional radar baselines. However, radar sensors can suffer from interference and clutter, particularly in urban areas or heavily congested electromagnetic environments \cite{Wu2024multi}.

\noindent\textbf{(2) Vision-Based Detection.}~Vision-based techniques remain widely adopted, benefiting from readily available cameras and sophisticated CNN-based detectors \cite{zhu2020vision, tan2020efficientdet}. Early work centered on region-based methods \cite{ren2016faster} and YOLO architectures \cite{bochkovskiy2020yolov4, wang2022yolov7}, gradually evolving into more lightweight or attention-oriented variants \cite{Cheng2024LocalGlobal, huang2024EDGSYOLOv8, hao2025yolov8}. While mosaic augmentation and specialized modules can improve small-object detection \cite{Dadboud2021yolov5, Cheng2024LocalGlobal}, vision-based approaches remain vulnerable to occlusions, clutter, and extreme distances. Researchers have thus explored thermal and infrared (IR) imaging \cite{Sun2024multiyolov8, Nair2024ThermalFL} or speed-accuracy trade-offs for real-time embedded scenarios \cite{bo2024yolov7gs, Elsayed2024LERFNetAE}.  

\noindent\textbf{(3) Multi-Modal Fusion.}~Single-modality approaches often fail under suboptimal conditions—low-visibility (vision), interference (radar), or noise (audio) \cite{al2019droneaudio, uddin2022tvddt}. Consequently, multi-modal fusion has gained traction as a means of leveraging complementary sensor data \cite{svanstrom2021dataset, arsenos2024}. Svanstr\"om et al.~\cite{svanstrom2021dataset} showed that augmenting YOLO-based RGB or thermal imaging with audio features increased F1 scores and reduced false alarms. Similarly, Wu et al.~\cite{Wu2024multi} fused radar and RGB streams via attention-based spatiotemporal networks to improve occlusion handling. Practical adoption, however, demands meticulous sensor synchronization, balanced hardware costs, robust computational efficiency, and real-time data association \cite{Rebbapragada2024C2FDrone}.

\noindent\textbf{Key Insights \& Challenges.}~From prior work \cite{wang2024survey, zhu2020vision, tan2020efficientdet}, several recurring themes emerge. 
\emph{1) Small UAV Detection:} Limited pixel coverage and subtle micro-Doppler signatures complicate detection \cite{He2024FogUAV, carion2020end, liu2021swin}, prompting multi-scale feature extraction, deformable convolutions, and transformer backbones \cite{tan2020efficientdet, Cheng2024LocalGlobal, huang2024EDGSYOLOv8}. Ensuring real-time operation at high accuracy remains challenging \cite{wang2024survey, wang2022yolov7}. 
\emph{2) Environmental Factors:} Ambient noise (audio), lighting (vision), and radio interference (radar) degrade performance \cite{al2019droneaudio, uddin2022tvddt, singh2024VisionUAV, zhao2021mdrone}, highlighting the need for advanced clutter suppression and robust sensor hardware. 
\emph{3) Generalization vs. Specialization:} YOLO-based models offer generalizable frameworks \cite{Dadboud2021yolov5, Cheng2024LocalGlobal, hao2025yolov8}, but specialized domains (e.g., stealth UAVs) may require niche architectures or domain-adaptive strategies \cite{Wu2024multi, bo2024yolov7gs}. 
\emph{4) Multi-Modal Fusion:} Complementary sensors effectively reduce false positives and enhance accuracy \cite{svanstrom2021dataset, Wu2024multi}, although issues like sensor alignment, real-time latency, and hardware cost must be resolved \cite{arsenos2024, Rebbapragada2024C2FDrone}. 
\emph{5) Benchmarking \& Deployment:} Heterogeneous data sources, labeling formats, and test conditions \cite{du2018unmanned, Paweczyk2020RealWO, zhu2020vision} complicate standardized evaluations. Though laboratory demonstrations show promise, real-time performance in crowded outdoor settings remains largely underexplored \cite{Fan2021DesignAI, Filkin2021UnmannedAV}. Overcoming these challenges will be crucial for field-ready anti-UAV systems.

%% file: sec/fig-2.tex
\tikzstyle{my-box}=[
    rectangle,
    draw=gray,
    rounded corners,
    text opacity=1,
    minimum height=1.5em,
    minimum width=5em,
    inner sep=2pt,
    align=center,
    fill opacity=.5,
    line width=0.8pt,
]
\tikzstyle{leaf}=[my-box, minimum height=1.5em, 
    fill=pink!10, text=black, align=left, font=\normalsize,
    inner xsep=2pt,
    inner ysep=4pt,
    line width=0.8pt,
]

\definecolor{c1}{RGB}{93,191,237} 
\definecolor{c2}{RGB}{237,110,106} 
\definecolor{c3}{RGB}{240,154,69}  
\definecolor{c4}{RGB}{108,222,157} 
\definecolor{c5}{RGB}{205,180,243} 
\definecolor{c6}{RGB}{97,218,184} 
\definecolor{c7}{RGB}{226,115,150} 
\definecolor{c8}{RGB}{201,116,201} 
\definecolor{c9}{RGB}{23,182,179} 
\definecolor{c10}{RGB}{242,157,108} 

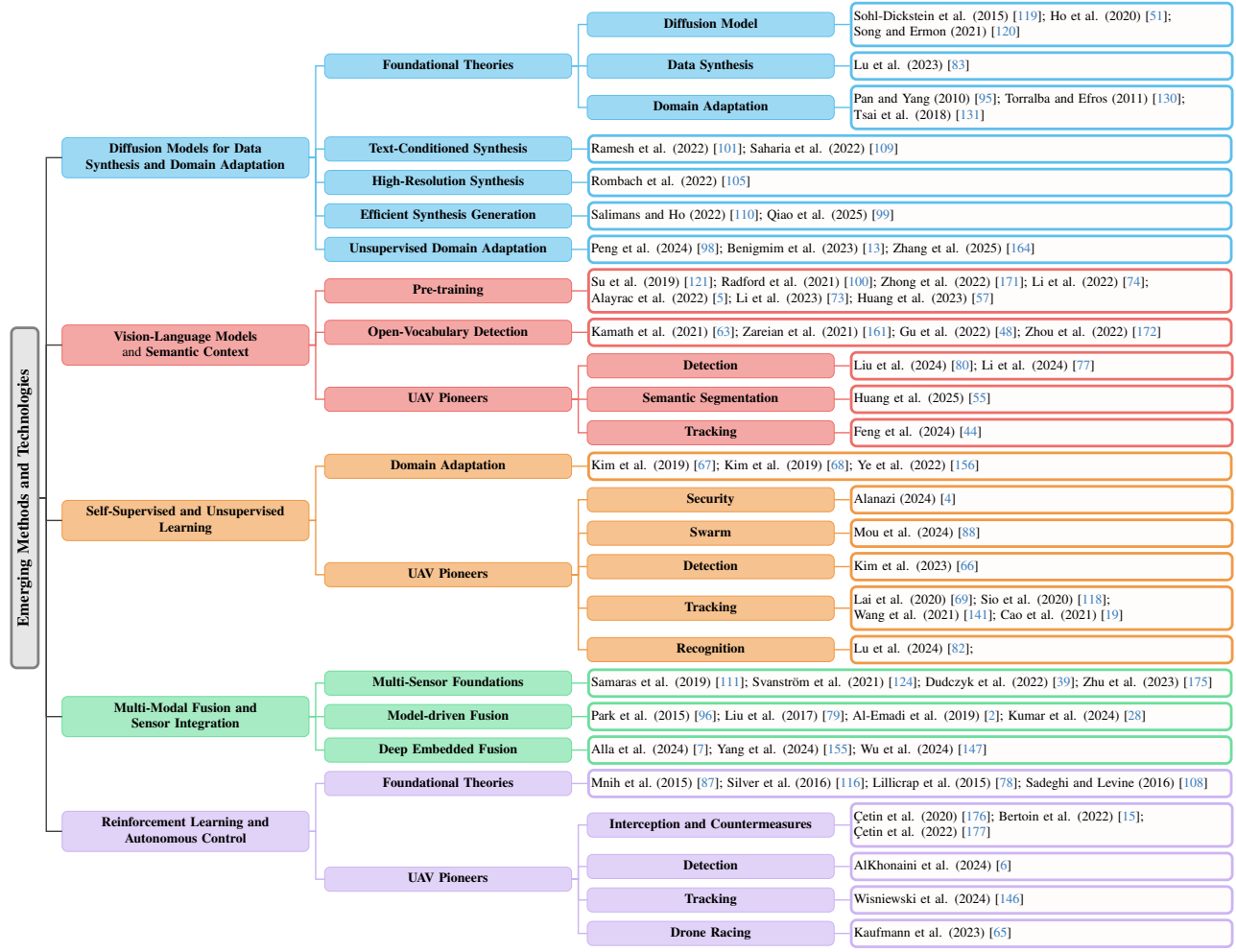
\begin{figure*}[!t]
    \centering
    \resizebox{0.97\textwidth}{!}{
        \begin{forest}
            forked edges,
            for tree={
                grow=east,
                reversed=true,
                anchor=base west,
                parent anchor=east,
                child anchor=west,
                base=center,
                font=\large,
                rectangle,
                draw=gray,
                rounded corners,
                align=left,
                text centered,
                minimum width=4em,
                edge+={darkgray, line width=1pt},
                s sep=3pt,
                inner xsep=2pt,
                inner ysep=3pt,
                line width=0.8pt,
                ver/.style={rotate=90, child anchor=north, parent anchor=south, anchor=center},
            },
            where level=1{text width=18em,font=\normalsize,}{},
            where level=2{text width=18em,font=\normalsize,}{},
            where level=3{text width=18em,font=\normalsize,}{},
            where level=4{text width=35em,font=\normalsize,}{},
            where level=5{text width=12em,font=\normalsize,}{},
            [
                \textbf{Emerging Methods and Technologies}, ver, line width=0.7mm,fill=gray!20, text width=25em
                [
                \begin{tabular}{c}\textbf{Diffusion Models for Data}\\\textbf{Synthesis and Domain Adaptation} \end{tabular}, fill=c1!60, draw=c1, line width=0mm
                    [
                    \textbf{Foundational Theories}, fill=c1!60, draw=c1, line width=0mm, edge={c1}
                        [
                        \textbf{Diffusion Model},fill=c1!60, draw=c1, line width=0mm, edge={c1}
                            [
                            Sohl-Dickstein et al. (2015) \cite{sohl2015deep}; Ho et al. (2020) \cite{ho2020denoising}; \\Song and Ermon (2021) \cite{song2021score}, leaf, draw=c1, line width=0.7mm, text width=28em, edge={c1}
                            ]  
                        ]
                        [
                        \textbf{Data Synthesis},fill=c1!60, draw=c1, line width=0mm, edge={c1}
                            [
                            Lu et al. (2023) \cite{lu2023machine}, leaf, draw=c1, line width=0.7mm, text width=28em, edge={c1}         
                            ]
                        ]
                        [
                        \textbf{Domain Adaptation},fill=c1!60, draw=c1, line width=0mm,edge={c1}
                            [
                            Pan and Yang (2010) \cite{pan2010survey}; Torralba and Efros (2011) \cite{torralba2011unbiased}; \\Tsai et al. (2018) \cite{tsai2018learning}, leaf, draw=c1, line width=0.7mm, text width=28em, edge={c1}
                            ] 
                        ]
                    ]
                    [
                    \textbf{Text-Conditioned Synthesis}, fill=c1!60, draw=c1, line width=0mm,edge={c1}
                            [
                            Ramesh et al. (2022) \cite{ramesh2022dalle2}; Saharia et al. (2022) \cite{saharia2022imagen}, leaf, draw=c1, line width=0.7mm, text width=47.5em, edge={c1}
                            ]
                    ]
                    [
                    \textbf{High-Resolution Synthesis}, fill=c1!60, draw=c1, line width=0mm,edge={c1}
                            [
                            Rombach et al. (2022) \cite{rombach2022highresolution}, leaf, draw=c1, line width=0.7mm, text width=47.5em, edge={c1}
                            ]
                    ]
                    [
                    \textbf{Efficient Synthesis Generation}, fill=c1!60, draw=c1, line width=0mm,edge={c1}
                            [
                            Salimans and Ho (2022) \cite{salimans2022progressive}; Qiao et al. (2025) \cite{qiao2025fast}, leaf, draw=c1, line width=0.7mm, text width=47.5em, edge={c1}
                            ]
                    ]
                    [
                    \textbf{Unsupervised Domain Adaptation}, fill=c1!60, draw=c1, line width=0mm, edge={c1}
                            [
                            Peng et al. (2024) \cite{peng2024unsupervised}; Benigmim et al. (2023) \cite{benigmim2023one}; Zhang et al. (2025) \cite{zhang2025domain}, leaf, draw=c1, line width=0.7mm, text width=47.5em, edge={c1}
                            ]
                    ]
                ]
                [
                \begin{tabular}{c} \textbf{Vision-Language Models} \\and \textbf{Semantic Context} \end{tabular}, fill=c2!60, draw=c2, line width=0mm
                    [
                    \textbf{Pre-training}, fill=c2!60, draw=c2, line width=0mm, edge={c2}
                        [
                        Su et al. (2019) \cite{Su2019VLBERTPO}; Radford et al. (2021) \cite{Radford2021LearningTV}; Zhong et al. (2022) \cite{Zhong_2022_CVPR}; Li et al. (2022) \cite{Li_2022_CVPR}; \\ Alayrac et al. (2022) \cite{Alayrac2022FlamingoAV}; Li et al. (2023) \cite{Li2023BLIP2BL}; Huang et al. (2023) \cite{Huang2023LanguageIN}, leaf, draw=c2, line width=0.7mm, text width=47.5em, edge={c2}
                        ]
                    ]
                    [
                    \textbf{Open-Vocabulary Detection}, fill=c2!60, draw=c2, line width=0mm, edge={c2}
                        [
                         Kamath et al. (2021) \cite{Kamath_2021_ICCV}; Zareian et al. (2021) \cite{Zareian_2021_CVPR}; Gu et al. (2022) \cite{Gu_2022_ICLR}; Zhou et al. (2022) \cite{Zhou_2022_ECCV}, leaf, draw=c2, line width=0.7mm, text width=47.5em, edge={c2}
                        ]
                    ]
                    [
                    \textbf{UAV Pioneers},  fill=c2!60, draw=c2, line width=0mm, edge={c2}
                        [
                        \textbf{Detection}, fill=c2!60, draw=c2, line width=0mm, edge={c2}
                            [
                            Liu et al. (2024) \cite{Liu_2024_ESWA}; Li et al. (2024) \cite{Li_2024_ECCV}, leaf, draw=c2, line width=0.7mm, text width=28em, edge={c2}
                            ]
                        ]
                        [
                        \textbf{Semantic Segmentation}, fill=c2!60, draw=c2, line width=0mm, edge={c2}
                            [
                            Huang et al. (2025) \cite{Huang_2025_Drones}, leaf, draw=c2, line width=0.7mm, text width=28em, edge={c2}
                            ]
                        ]
                        [
                        \textbf{Tracking}, fill=c2!60, draw=c2, line width=0mm, edge={c2}
                            [
                            Feng et al. (2024) \cite{Feng_2024_NeurIPS}, leaf, draw=c2, line width=0.7mm, text width=28em, edge={c2}
                            ]
                        ]
                    ]
                ]
                [
                    \begin{tabular}{c} \textbf{Self-Supervised and Unsupervised} \\\textbf{Learning} \end{tabular}, fill=c3!60, draw=c3, line width=0mm
                    [
                    \textbf{Domain Adaptation}, fill=c3!60, draw=c3, line width=0mm, edge={c3}
                        [
                        Kim et al. (2019) \cite{Kim_2019_ICCV}; Kim et al. (2019) \cite{Kim_2019_CVPR}; Ye et al. (2022) \cite{Ye_2022_CVPR}, leaf, draw=c3, line width=0.7mm, text width=47.5em, edge={c3}
                        ]
                    ]
                    [
                    \textbf{UAV Pioneers},  fill=c3!60, draw=c3, line width=0mm, edge={c3}
                        [
                        \textbf{Security}, fill=c3!60, draw=c3, line width=0mm, edge={c3}
                            [
                            Alanazi (2024) \cite{Alanazi2024SSRL}, leaf, draw=c3, line width=0.7mm, text width=28em, edge={c3}
                            ]
                        ]
                        [
                        \textbf{Swarm}, fill=c3!60, draw=c3, line width=0mm, edge={c3}
                            [
                            Mou et al. (2024) \cite{Mou2024luster}, leaf, draw=c3, line width=0.7mm, text width=28em, edge={c3}
                            ]
                        ]
                        [
                        \textbf{Detection}, fill=c3!60, draw=c3, line width=0mm, edge={c3}
                            [
                            Kim et al. (2023) \cite{10473588}, leaf, draw=c3, line width=0.7mm, text width=28em, edge={c3}
                            ]
                        ]
                        [
                        \textbf{Tracking}, fill=c3!60, draw=c3, line width=0mm, edge={c3}
                            [
                             Lai et al. (2020) \cite{Lai_2020_CVPR}; Sio et al. (2020) \cite{Sio_2020_MM};\\ Wang et al. (2021) \cite{Wang_2021_IJCV}; Cao et al. (2021) \cite{Cao_2021_ICCV}, leaf, draw=c3, line width=0.7mm, text width=28em, edge={c3}
                            ]
                        ]
                        [
                        \textbf{Recognition}, fill=c3!60, draw=c3, line width=0mm, edge={c3}
                            [
                            Lu et al. (2024) \cite{10683145};, leaf, draw=c3, line width=0.7mm, text width=28em, edge={c3}
                            ]
                        ]
                    ]
                ]
                [
                \begin{tabular}{c} \textbf{Multi-Modal Fusion and} \\\textbf{Sensor Integration} \end{tabular}, fill=c4!60, draw=c4, line width=0mm
                    [
                    \textbf{Multi-Sensor Foundations}, fill=c4!60, draw=c4, line width=0mm, edge={c4}
                        [
                        Samaras et al. (2019) \cite{samaras2019multisensor}; Svanström et al. (2021) \cite{svanstrom2021dataset}; Dudczyk et al. (2022) \cite{dudczyk2022multisensory}; Zhu et al. (2023) \cite{anti_uav_600}, leaf, draw=c4, line width=0.7mm, text width=47.5em, edge={c4}
                        ]
                    ]
                    [
                    \textbf{Model-driven Fusion}, fill=c4!60, draw=c4, line width=0mm, edge={c4}
                        [
                        Park et al. (2015) \cite{park2015radarAudio}; Liu et al. (2017) \cite{liu2017audioCamera}; Al-Emadi et al. (2019) \cite{al2019droneaudio}; Kumar et al. (2024) \cite{kumar2024radarCamera}, leaf, draw=c4, line width=0.7mm, text width=47.5em, edge={c4}
                        ]
                    ]
                    [
                    \textbf{Deep Embedded Fusion}, fill=c4!60, draw=c4, line width=0mm, edge={c4}
                        [
                        Alla et al. (2024) \cite{Alla2024}; Yang et al. (2024) \cite{Yang2024AVFDTI}; Wu et al. (2024) \cite{Wu2024multi}, leaf, draw=c4, line width=0.7mm, text width=47.5em, edge={c4}
                        ]
                    ]
                ]
                [
                \begin{tabular}{c} \textbf{Reinforcement Learning and} \\\textbf{Autonomous Control} \end{tabular}, fill=c5!60, draw=c5, line width=0mm
                    [
                    \textbf{Foundational Theories}, fill=c5!60, draw=c5, line width=0mm, edge={c5}
                        [
                        Mnih et al.~(2015)~\cite{mnih2015human}; Silver et al.~(2016)~\cite{silver2016mastering}; Lillicrap et al.~(2015)~\cite{lillicrap2015continuous}; Sadeghi and Levine~(2016)~\cite{sadeghi2016cad2rl}, leaf, draw=c5, line width=0.7mm, text width=47.5em, edge={c5}
                        ]
                    ]
                    [
                    \textbf{UAV Pioneers}, fill=c5!60, draw=c5, line width=0mm, edge={c5}
                        [
                        \textbf{Interception and Countermeasures}, fill=c5!60, draw=c5, line width=0mm, edge={c5}
                            [
                            Çetin et al. (2020) \cite{cetin2020}; Bertoin et al. (2022)   \cite{bertoin2022autonomous};\\ Çetin et al. (2022) \cite{cetin2022}, leaf, draw=c5, line width=0.7mm, text width=28em, edge={c5}
                            ]
                        ]
                        [
                        \textbf{Detection}, fill=c5!60, draw=c5, line width=0mm, edge={c5}
                            [
                            AlKhonaini et al. (2024) \cite{alkhonaini2024}, leaf, draw=c5, line width=0.7mm, text width=28em, edge={c5}
                            ]
                        ]
                        [
                        \textbf{Tracking}, fill=c5!60, draw=c5, line width=0mm, edge={c5}
                            [
                            Wisniewski et al. (2024) \cite{wisniewski2024towards}, leaf, draw=c5, line width=0.7mm, text width=28em, edge={c5}
                            ]
                        ]
                        [
                        \textbf{Drone Racing}, fill=c5!60, draw=c5, line width=0mm, edge={c5}
                            [
                            Kaufmann et al. (2023) \cite{kaufmann2023champion}, leaf, draw=c5, line width=0.7mm, text width=28em, edge={c5}
                            ]
                        ]
                    ]
                ]
            ]
        \end{forest}
    }
    \caption{\small Taxonomy for emerging methods and technologies in UAV research.}
    \label{fig:taxonomy}
    \vspace{-0.2in}
\end{figure*}

%% file: sec/5-tracking.tex
\section{Anti-UAV Tracking Approaches} 
\label{sec:track_survey}

Table~\ref{tab:final_table_interpretations_tracking} reviews prominent UAV tracking methods spanning detection-based, Siamese-based, and hybrid trackers, employing RGB, IR, and multi-modal inputs. Although direct numerical comparisons are hampered by dataset and metric variations, these approaches collectively represent the state of the art in UAV tracking. 

\noindent\textbf{(1) Detection-Based Trackers.} Detection-based pipelines usually prioritize real-time performance and utilize efficient object detectors. For example, Wu et al.\ \cite{Wu2024EagleEye} proposed EECF, a dual-fovea correlation filter tracker operating at 86\,FPS with a state accuracy score of 66.42\% on Anti-UAV \cite{anti_uav} and 70.45\% on USC-Drone \cite{USC_Drone}. Wang et al.\ \cite{wang2024target} integrated YOLOX \cite{YOLOX} with Swin Transformer \cite{liu2021swin} and ByteTrack, achieving 78.2\% accuracy on UAVSwarm \cite{wang2022uavswarm}. Deng et al.\ \cite{deng2024multi} combined YOLOv9 \cite{Wang2024YOLOv9LW} and EfficientNet-B7 \cite{Tan2019EfficientNetRM} with LSTM/Kalman filtering for 3D tracking, earning top-ranked pose MSE on MMAUD \cite{yuan2024mmaud} and winning the UG2+ Challenge \cite{UG2Challenge2024}.

\noindent\textbf{(2) Siamese (Contrastive) Trackers.} Siamese trackers excel in learning robust feature similarities and handling fast or evasive UAV motion. Huang et al.\ \cite{Huang2021SiamSTA} claimed first-place in the 1st and 2nd Anti-UAV Challenges \cite{anti_uav, zhao20212nd} with SiamSTA, which incorporates sophisticated spatio-temporal attention and motion estimation. Li et al.\ \cite{li2022dual} augmented SuperDiMP with a dual-branch IR fusion strategy, boosting precision by 4.52\% in the 2nd Anti-UAV Challenge. Chen et al.\ \cite{Chen2022SiamSTA} further enhanced SiamSTA with CDCF and a multi-stage re-detection mechanism, excelling against small, fast UAVs. Beyond single-modality, Zhang et al.\ \cite{zhang2023modality}’s SiamFusion seamlessly integrated RGB and IR streams for a 63.37 score on Anti-UAV, illustrating the tangible value of complementary sensor modalities.

\noindent\textbf{(3) Hybrid (Detection + Siamese) Trackers.}~Hybrid methods combine detection pipelines with Siamese matching, capitalizing on each paradigm's respective strengths. Cheng et al.\ \cite{Cheng2022AntiUAVTracking} introduced SiamAD, a synthesis of YOLOv5, multi-hybrid attention modules, and SiamRPN++, achieving 88.4\% detection precision at 37.1\,FPS on the Anti-UAV dataset \cite{anti_uav}. Yu et al.\ \cite{Yu2023UnifiedTransformer} integrated Transformers, YOLOv5, LoFTR, GMM, and OSTrack for second place in the 3rd Anti-UAV Challenge \cite{zhao20233rd}, affirming the practical effectiveness of multi-stage IR-based tracking.

\noindent\textbf{Key Insights \& Challenges.}~Although real-time Siamese trackers \cite{shen2022real} and detection-based pipelines \cite{Fang2021Real} can sustain high frame rates, they often struggle under clutter, occlusion, or swarm conditions, thereby motivating more adaptive architectures \cite{yan2021learning, chen2021transformer}. Stealth maneuvers and jamming further complicate UAV tracking, emphasizing the necessity for robust domain adaptation and dynamic cue prioritization \cite{do2025ramots}. While RGBT sensor fusion \cite{Li2019Multi,tang2024revisiting} mitigates individual sensor shortcomings, incorporating LiDAR or radar \cite{deng2024multi} inevitably increases calibration overhead and computational load. Existing benchmarks rarely capture large-scale swarms, diverse weather conditions, or severe occlusions, consequently restricting methodical progress \cite{Wu2021Survey}. Transformer-based solutions \cite{chen2021transformer,cui2022mixformer,xiao2024target} address complex motion patterns and partial occlusion but can lag in real-time performance, necessitating specialized hardware optimizations for on-board or edge devices. Extending multi-object trackers (e.g., ByteTrack \cite{zhang2022bytetrack}) to larger UAV swarms remains challenging due to association uncertainty and domain-specific constraints. Ultimately, next-generation trackers will demand faster inference, improved sensor fusion, and broader datasets to achieve reliable anti-UAV performance in real-world conditions.

%% file: sec/6-emerging_future.tex
\section{Emerging Techniques \& Future Directions}
\label{sec:future}

Recent UAV research has embraced advanced machine learning paradigms to boost performance, scalability, and autonomy. Figure~\ref{fig:taxonomy} categorizes these innovations into five key themes: (\emph{i}) diffusion models for data synthesis and domain adaptation, (\emph{ii}) vision-language models for semantic understanding, (\emph{iii}) self-/unsupervised learning, (\emph{iv}) multi-modal fusion and sensor integration, and (\emph{v}) reinforcement learning for autonomous control. We outline critical developments and future opportunities below.

\noindent\textbf{Diffusion Models for Data Synthesis \& Domain Adaptation.}
Diffusion models \cite{ho2020denoising,song2021score} generate high-fidelity UAV imagery to address data scarcity and enhance domain diversity, including text-conditioned approaches \cite{ramesh2022dalle2,saharia2022imagen}. Recent advances in faster sampling \cite{salimans2022progressive,qiao2025fast} support near real-time augmentation, while diffusion-based domain adaptation \cite{peng2024unsupervised,benigmim2023one,zhang2025domain} narrows the gap between synthetic and real-world data. Future research may focus on on-the-fly adaptation from live sensor streams, enabling continuous refinement under dynamic UAV operating conditions.

\noindent\textbf{Vision-Language Models \& Semantic Context.}
Vision-language (VL) methods integrate textual cues with visual features to facilitate open-vocabulary UAV detection and classification \cite{Radford2021LearningTV,Zhong_2022_CVPR,Zareian_2021_CVPR,li2024human,dong2025ha, dong2025unified}. UAV-specific adaptations \cite{Liu_2024_ESWA,Huang_2025_Drones,Li_2024_ECCV} incorporate textual descriptions on flight profiles, mission goals, or payload attributes, thereby \emph{significantly} enriching threat assessment. By leveraging semantic narratives~\cite{cheng2024shield} (e.g., infiltration routes, suspect cargo), future VL-based systems can \emph{swiftly} adapt to novel UAV categories and enhance situational awareness.

\noindent\textbf{Self-Supervised \& Unsupervised Learning.}
Self- and unsupervised approaches alleviate dependence on labeled aerial data across varying altitudes, sensors, and weather \cite{Kim_2019_ICCV,Kim_2019_CVPR,Ye_2022_CVPR}. Recent work addresses swarm-wide generalization \cite{Mou2024luster}, \emph{further} strengthening multi-UAV detection \cite{10473588}, tracking \cite{Wang_2021_IJCV,Lai_2020_CVPR}, and recognition \cite{Sio_2020_MM}, as well as security \cite{Alanazi2024SSRL}. Moving forward, large-scale self-supervised pre-training that \emph{rapidly} adapts to domain shifts in real time may enable \emph{ongoing} responses to evolving UAV behaviors and stealth tactics.

\noindent\textbf{Multi-Modal Fusion \& Sensor Integration.}
Combining sensors (e.g., RGB, IR, radar, and audio) \emph{significantly} boosts UAV detection and tracking robustness \cite{park2015radarAudio,messina2019radar,samaras2019multisensor,anti_uav_600}. Classical fusion \cite{park2015radarAudio,liu2017audioCamera} is now augmented by deep embedding methods \cite{Alla2024,Yang2024AVFDTI}, allowing adaptive sensor weighting under real-time constraints. However, \emph{maintaining} precise synchronization \cite{Wu2024multi} and carefully balancing power or bandwidth remain essential considerations for \emph{robust}, large-scale deployments.

\noindent\textbf{Reinforcement Learning \& Autonomous Control.}
Reinforcement Learning (RL) has proven effective for adaptive UAV decision-making in dynamic environments, building on foundational breakthroughs \cite{mnih2015human,silver2016mastering}. Modern UAV-centric RL research tackles interception \cite{bertoin2022autonomous,cetin2022}, detection \cite{alkhonaini2024}, tracking \cite{wisniewski2024towards}, and high-speed racing \cite{kaufmann2023champion}, frequently integrating generative or multi-modal inputs for rapid policy updates. Future directions may emphasize hierarchical or multi-agent RL for large-scale swarm control, balancing cooperative behavior with adversarial countermeasures and mission-specific constraints.

%% file: sec/9-conclusion.tex
\section{Conclusion}
\label{sec:conclusion}
This survey examined the evolving anti-UAV landscape, focusing on detection, tracking, and recognition. We highlighted recent progress in sensor fusion, deep learning, and multi-modal systems, emphasizing the urgency of robust, real-time solutions under adverse conditions. Analysis of representative datasets revealed gaps in adversarial robustness and scalable swarm tracking. Key insights underscore cross-modal integration, resilience to adversarial tactics, and the promise of self-supervised and reinforcement learning. Moving forward, generative modeling, semantic understanding, and adaptive multi-modal frameworks are poised to drive the next wave of anti-UAV innovations.

%% file: main.bib
@String(CVPR= {IEEE Conf. Comput. Vis. Pattern Recog.})

@String(ICCV= {Int. Conf. Comput. Vis.})

@String(ECCV= {Eur. Conf. Comput. Vis.})

@String(NIPS= {Adv. Neural Inform. Process. Syst.})

@String(ICPR = {Int. Conf. Pattern Recog.})

@String(ICASSP=	{ICASSP})

@String(ICIP = {IEEE Int. Conf. Image Process.})

@String(ICLR = {Int. Conf. Learn. Represent.})

@String(CVPRW= {IEEE Conf. Comput. Vis. Pattern Recog. Worksh.})

@String(CVPR  = {CVPR})

@String(ICCV  = {ICCV})

@String(ECCV  = {ECCV})

@String(NIPS  = {NeurIPS})

@String(ICPR  = {ICPR})

@String(ICIP  = {ICIP})

@String(ICLR  = {ICLR})

@String(CVPRW= {CVPRW})

@INPROCEEDINGS{rebbapragada2024c2fdrone,
  author={Rebbapragada, Sairam VC and Panda, Pranoy and Balasubramanian, Vineeth N},
  booktitle={2024 IEEE International Conference on Robotics and Automation (ICRA)}, 
  title={C2FDrone: Coarse-to-Fine Drone-to-Drone Detection using Vision Transformer Networks}, 
  year={2024},
  volume={},
  number={},
  pages={6627-6633},
  doi={10.1109/ICRA57147.2024.10609997}}

@INPROCEEDINGS{li2016multi,
  author={Li, Jing and Ye, Dong Hye and Chung, Timothy and Kolsch, Mathias and Wachs, Juan and Bouman, Charles},
  booktitle={2016 IEEE/RSJ International Conference on Intelligent Robots and Systems (IROS)}, 
  title={Multi-target detection and tracking from a single camera in Unmanned Aerial Vehicles (UAVs)}, 
  year={2016},
  volume={},
  number={},
  pages={4992-4997},
  doi={10.1109/IROS.2016.7759733}}

@ARTICLE{rozantsev2016detecting,
  author={Rozantsev, Artem and Lepetit, Vincent and Fua, Pascal},
  journal={IEEE Transactions on Pattern Analysis and Machine Intelligence}, 
  title={Detecting Flying Objects Using a Single Moving Camera}, 
  year={2017},
  volume={39},
  number={5},
  pages={879-892},
  doi={10.1109/TPAMI.2016.2564408}}

@INPROCEEDINGS{AlEmadi2020cnn,
  author={Al-Emadi, Sara and Al-Senaid, Felwa},
  booktitle={2020 IEEE International Conference on Informatics, IoT, and Enabling Technologies (ICIoT)}, 
  title={Drone Detection Approach Based on Radio-Frequency Using Convolutional Neural Network}, 
  year={2020},
  volume={},
  number={},
  pages={29-34},
  doi={10.1109/ICIoT48696.2020.9089489}
}

@INPROCEEDINGS{al2019droneaudio,
  author={Al-Emadi, Sara and Al-Ali, Abdulla and Mohammad, Amr and Al-Ali, Abdulaziz},
  booktitle={2019 15th International Wireless Communications \& Mobile Computing Conference (IWCMC)}, 
  title={Audio Based Drone Detection and Identification using Deep Learning}, 
  year={2019},
  volume={},
  number={},
  pages={459-464},
  doi={10.1109/IWCMC.2019.8766732}}

@article{AlSaad2019,
  title={RF-based drone detection and identification using deep learning approaches: An initiative towards a large open source drone database},
  author={Al-Sa’d, Mohammad F and Al-Ali, Abdulla and Mohamed, Amr and Khattab, Tamer and Erbad, Aiman},
  journal={Future Generation Computer Systems},
  volume={100},
  pages={86--97},
  year={2019},
  publisher={Elsevier}
}

@Article{Alanazi2024SSRL,
AUTHOR = {Alanazi, Abed},
TITLE = {SSRL-UAVs: A Self-Supervised Deep Representation Learning Approach for GPS Spoofing Attack Detection in Small Unmanned Aerial Vehicles},
JOURNAL = {Drones},
VOLUME = {8},
YEAR = {2024},
NUMBER = {9},
ARTICLE-NUMBER = {515}
}

@article{Alayrac2022FlamingoAV,
  title={Flamingo: a visual language model for few-shot learning},
  author={Alayrac, Jean-Baptiste and Donahue, Jeff and Luc, Pauline and Miech, Antoine and Barr, Iain and Hasson, Yana and Lenc, Karel and Mensch, Arthur and Millican, Katherine and Reynolds, Malcolm and others},
  journal={Advances in neural information processing systems},
  volume={35},
  pages={23716--23736},
  year={2022}
}

@Article{alkhonaini2024,
    AUTHOR = {AlKhonaini, Arwa and Sheltami, Tarek and Mahmoud, Ashraf and Imam, Muhammad},
    TITLE = {UAV Detection Using Reinforcement Learning},
    JOURNAL = {Sensors},
    VOLUME = {24},
    YEAR = {2024},
    NUMBER = {6},
    ARTICLE-NUMBER = {1870},
    URL = {https://www.mdpi.com/1424-8220/24/6/1870},
    PubMedID = {38544131},
    ISSN = {1424-8220},
    DOI = {10.3390/s24061870}
}

@inproceedings{Alla2024,
    author = {Alla, Ildi and Olou, Herv\'{e} B. and Loscri, Valeria and Levorato, Marco},
    title = {From Sound to Sight: Audio-Visual Fusion and Deep Learning for Drone Detection},
    year = {2024},
    isbn = {9798400705823},
    publisher = {Association for Computing Machinery},
    address = {New York, NY, USA},
    url = {https://doi.org/10.1145/3643833.3656133},
    doi = {10.1145/3643833.3656133},
    booktitle = {Proceedings of the 17th ACM Conference on Security and Privacy in Wireless and Mobile Networks},
    pages = {123–133},
    numpages = {11},
    location = {Seoul, Republic of Korea},
    series = {WiSec '24}
}

@article{allahham2019dronerf,
  title={DroneRF dataset: A dataset of drones for RF-based detection, classification and identification},
  author={Allahham, MHD Saria and Al-Sa'd, Mohammad F and Al-Ali, Abdulla and Mohamed, Amr and Khattab, Tamer and Erbad, Aiman},
  journal={Data in brief},
  volume={26},
  pages={104313},
  year={2019}
}

@inproceedings{Allahham2020,
  title={Deep learning for RF-based drone detection and identification: A multi-channel 1-D convolutional neural networks approach},
  author={Allahham, Mhd Saria and Khattab, Tamer and Mohamed, Amr},
  booktitle={2020 IEEE International Conference on Informatics, IoT, and Enabling Technologies (ICIoT)},
  pages={112--117},
  year={2020},
  organization={IEEE}
}

@article{peng2024unsupervised,
  title={Unsupervised domain adaptation via domain-adaptive diffusion},
  author={Peng, Duo and Ke, Qiuhong and Ambikapathi, ArulMurugan and Yazici, Yasin and Lei, Yinjie and Liu, Jun},
  journal={IEEE Transactions on Image Processing},
  year={2024},
  publisher={IEEE}
}

@article{Bernardini2017,
  title={Drone detection by acoustic signature identification},
  author={Bernardini, Andrea and Mangiatordi, Federica and Pallotti, Emiliano and Capodiferro, Licia},
  journal={electronic imaging},
  volume={29},
  pages={60--64},
  year={2017},
  publisher={Society for Imaging Science and Technology}
}

@inproceedings{bertoin2022autonomous,
  title={Autonomous drone interception with deep reinforcement learning},
  author={Bertoin, David and Gauffriau, Adrien and Grasset, Damien and Gupta, Jayant Sen},
  booktitle={12th International Workshop on Agents in Traffic and Transportation (ATT 2022) in conjunction with IJCAI-ECAI 2022},
  number={3173},
  year={2022}
}

@Article{bo2024yolov7gs,
AUTHOR = {Bo, Chunjuan and Wei, Yuntao and Wang, Xiujia and Shi, Zhan and Xiao, Ying},
TITLE = {Vision-Based Anti-UAV Detection Based on YOLOv7-GS in Complex Backgrounds},
JOURNAL = {Drones},
VOLUME = {8},
YEAR = {2024},
NUMBER = {7},
ARTICLE-NUMBER = {331},
URL = {https://www.mdpi.com/2504-446X/8/7/331},
ISSN = {2504-446X},
DOI = {10.3390/drones8070331}
}

@InProceedings{Cao_2021_ICCV,
  author    = {Cao, Ziang and Fu, Changhong and Ye, Junjie and Li, Bowen and Li, Yiming},
  title     = {HiFT: Hierarchical Feature Transformer for Aerial Tracking},
  booktitle = {Proceedings of the IEEE/CVF International Conference on Computer Vision (ICCV)},
  year      = {2021},
  pages     = {15457--15466}
}

@Article{2021Casabiancaacostic,
AUTHOR = {Casabianca, Pietro and Zhang, Yu},
TITLE = {Acoustic-Based UAV Detection Using Late Fusion of Deep Neural Networks},
JOURNAL = {Drones},
VOLUME = {5},
YEAR = {2021},
NUMBER = {3},
ARTICLE-NUMBER = {54},
URL = {https://www.mdpi.com/2504-446X/5/3/54},
ISSN = {2504-446X},
DOI = {10.3390/drones5030054}
}

@article{Chen2022SiamSTA,
  title     = {Learning Spatio-Temporal Attention Based Siamese Network for Tracking UAVs in the Wild},
  author    = {Junjie Chen and Bo Huang and Jianan Li and Ying Wang and Moxuan Ren and Tingfa Xu},
  journal   = {Remote Sensing},
  volume    = {14},
  number    = {8},
  pages     = {1797},
  year      = {2022},
  doi       = {10.3390/rs14081797},
  url       = {https://www.mdpi.com/2072-4292/14/8/1797}
}

@article{li2024human,
  title={Human-aware vision-and-language navigation: Bridging simulation to reality with dynamic human interactions},
  author={Li, Heng and Li, Minghan and Cheng, Zhi-Qi and Dong, Yifei and Zhou, Yuxuan and He, Jun-Yan and Dai, Qi and Mitamura, Teruko and Hauptmann, Alexander G},
  journal={Advances in Neural Information Processing Systems},
  volume={37},
  pages={119411--119442},
  year={2024}
}

@INPROCEEDINGS{zhang2023modality,
  author={Zhang, Zhihao and Jin, Lei and Li, Shengjie and Xia, JianQiang and Wang, Jun and Li, Zun and Zhu, Zheng and Yang, Wenhan and Zhang, PengFei and Zhao, Jian and Zhang, Bo},
  booktitle={2023 IEEE International Conference on Image Processing (ICIP)}, 
  title={Modality Meets Long-Term Tracker: A Siamese Dual Fusion Framework for Tracking UAV}, 
  year={2023},
  volume={},
  number={},
  pages={1975-1979},
  doi={10.1109/ICIP49359.2023.10222679}}

@article{xie2023stftrack,
  AUTHOR = {Xie, Xueli and Xi, Jianxiang and Yang, Xiaogang and Lu, Ruitao and Xia, Wenxin},
TITLE = {STFTrack: Spatio-Temporal-Focused Siamese Network for Infrared UAV Tracking},
JOURNAL = {Drones},
VOLUME = {7},
YEAR = {2023},
NUMBER = {5},
ARTICLE-NUMBER = {296},
URL = {https://www.mdpi.com/2504-446X/7/5/296},
ISSN = {2504-446X},
DOI = {10.3390/drones7050296}
}

@article{dong2025unified,
  title={Unified World Models: Memory-Augmented Planning and Foresight for Visual Navigation},
  author={Dong, Yifei and Wu, Fengyi and Chen, Guangyu and Cheng, Zhi-Qi and Hu, Qiyu and Zhou, Yuxuan and Sun, Jingdong and He, Jun-Yan and Dai, Qi and Hauptmann, Alexander G},
  journal={arXiv preprint arXiv:2510.08713},
  year={2025}
}

@misc{YOLOX,
      title={YOLOX: Exceeding YOLO Series in 2021}, 
      author={Zheng Ge and Songtao Liu and Feng Wang and Zeming Li and Jian Sun},
      year={2021},
      eprint={2107.08430},
      archivePrefix={arXiv},
      primaryClass={cs.CV},
      url={https://arxiv.org/abs/2107.08430}, 
}

@inproceedings{Wang2024YOLOv9LW,
  title={Yolov9: Learning what you want to learn using programmable gradient information},
  author={Wang, Chien-Yao and Yeh, I-Hau and Mark Liao, Hong-Yuan},
  booktitle={European conference on computer vision},
  pages={1--21},
  year={2024},
  organization={Springer}
}

@article{cheng2024shield,
  title={Shield: Llm-driven schema induction for predictive analytics in ev battery supply chain disruptions},
  author={Cheng, Zhi-Qi and Dong, Yifei and Shi, Aike and Liu, Wei and Hu, Yuzhi and O'Connor, Jason and Hauptmann, Alexander G and Whitefoot, Kate S},
  journal={arXiv preprint arXiv:2408.05357},
  year={2024}
}

@inproceedings{USC_Drone,
  title={A deep learning approach to drone monitoring},
  author={Chen, Yueru and Aggarwal, Pranav and Choi, Jongmoo and Kuo, C-C Jay},
  booktitle={2017 Asia-Pacific Signal and Information Processing Association Annual Summit and Conference (APSIPA ASC)},
  pages={686--691},
  year={2017},
  organization={IEEE}
}

@article{Cheng2022AntiUAVTracking,
  title     = {An Anti-UAV Long-Term Tracking Method with Hybrid Attention Mechanism and Hierarchical Discriminator},
  author    = {Feng Cheng and Zhiqiang Liang and Guoping Peng and Shun Liu and Shijie Li and Ming Ji},
  journal   = {Sensors},
  volume    = {22},
  number    = {10},
  pages     = {3701},
  year      = {2022},
  doi       = {10.3390/s22103701},
  url       = {https://www.mdpi.com/1424-8220/22/10/3701}
}

@INPROCEEDINGS{Cheng2024LocalGlobal,
  author={Cheng, Qian and Li, Jia and Du, Juan and Li, Shaojuan},
  booktitle={2024 IEEE 7th Advanced Information Technology, Electronic and Automation Control Conference (IAEAC)}, 
  title={Anti-UAV Detection Method Based on Local-Global Feature Focusing Module}, 
  year={2024},
  volume={7},
  number={},
  pages={1413-1418},
  doi={10.1109/IAEAC59436.2024.10503882}
}

@article{dong2025ha,
  title={HA-VLN: A Benchmark for Human-Aware Navigation in Discrete-Continuous Environments with Dynamic Multi-Human Interactions, Real-World Validation, and an Open Leaderboard},
  author={Dong, Yifei and Wu, Fengyi and He, Qi and Li, Heng and Li, Minghan and Cheng, Zebang and Zhou, Yuxuan and Sun, Jingdong and Dai, Qi and Cheng, Zhi-Qi and others},
  journal={arXiv preprint arXiv:2503.14229},
  year={2025}
}

@article{kumar2024radarCamera,
  title   = {{UAV} Detection Multi-sensor Data Fusion},
  author  = {Chiranjeevi, Amit Kumar and Giridhar, Ozkan},
  journal = {Journal of Research in Science and Engineering},
  volume  = {6},
  number  = {7},
  pages   = {6--12},
  year    = {2024},
  publisher = {Century Science Publishing Co},
  doi     = {10.53469/jrse.2024.06(07).02}
}

@inproceedings{chu2023experimental,
  title={An Experimental Evaluation Based on New Air-to-Air Multi-UAV Tracking Dataset},
  author={Chu, Zhaochen and Song, Tao and Jin, Ren and Jiang, Tao},
  booktitle={2023 IEEE International Conference on Unmanned Systems (ICUS)},
  pages={671--676},
  year={2023},
  organization={IEEE}
}

@inproceedings{Coluccia2021DroneBird,
  author    = {Angelo Coluccia and Andrea Fascista and Arne Schumann and Lukas Sommer and Anastasios Dimou and Dimitrios Zarpalas and et al.},
  title     = {Drone-vs-Bird Detection Challenge at {IEEE} AVSS 2021},
  booktitle = {17th IEEE International Conference on Advanced Video and Signal Based Surveillance (AVSS)},
  year      = {2021},
  pages     = {1--8},
  doi       = {10.1109/AVSS52988.2021.9625349}
}

@INPROCEEDINGS{Dadboud2021yolov5,
  author={Dadboud, Fardad and Patel, Vaibhav and Mehta, Varun and Bolic, Miodrag and Mantegh, Iraj},
  booktitle={2021 17th IEEE International Conference on Advanced Video and Signal Based Surveillance (AVSS)}, 
  title={Single-Stage UAV Detection and Classification with YOLOV5: Mosaic Data Augmentation and PANet}, 
  year={2021},
  volume={},
  number={},
  pages={1-8},
  doi={10.1109/AVSS52988.2021.9663841}
}

@misc{deng2024multi,
      title={Multi-Modal UAV Detection, Classification and Tracking Algorithm -- Technical Report for CVPR 2024 UG2 Challenge}, 
      author={Tianchen Deng and Yi Zhou and Wenhua Wu and Mingrui Li and Jingwei Huang and Shuhong Liu and Yanzeng Song and Hao Zuo and Yanbo Wang and Yutao Yue and Hesheng Wang and Weidong Chen},
      year={2024},
      eprint={2405.16464},
      archivePrefix={arXiv},
      primaryClass={cs.RO},
      url={https://arxiv.org/abs/2405.16464}, 
}

@article{dudczyk2022multisensory,
  title   = {Multi-Sensory Data Fusion in Terms of {UAV} Detection in 3D Space},
  author  = {Dudczyk, Janusz and Czyba, Roman and Skrzypczyk, Krzysztof},
  journal = {Sensors},
  volume  = {22},
  number  = {12},
  pages   = {4323},
  year    = {2022},
  publisher = {MDPI},
  doi     = {10.3390/s22124323}
}

@InProceedings{Fang2021Real,
    author    = {Fang, Houzhang and Wang, Xiaolin and Liao, Zikai and Chang, Yi and Yan, Luxin},
    title     = {A Real-Time Anti-Distractor Infrared UAV Tracker With Channel Feature Refinement Module},
    booktitle = {Proceedings of the IEEE/CVF International Conference on Computer Vision (ICCV) Workshops},
    month     = {October},
    year      = {2021},
    pages     = {1240-1248}
}

@Article{Frid2024,
AUTHOR = {Frid, Alan and Ben-Shimol, Yehuda and Manor, Erez and Greenberg, Shlomo},
TITLE = {Drones Detection Using a Fusion of RF and Acoustic Features and Deep Neural Networks},
JOURNAL = {Sensors},
VOLUME = {24},
YEAR = {2024},
NUMBER = {8},
ARTICLE-NUMBER = {2427},
URL = {https://www.mdpi.com/1424-8220/24/8/2427},
PubMedID = {38676050},
ISSN = {1424-8220},
DOI = {10.3390/s24082427}
}

@InProceedings{Gu_2022_ICLR,
  title     = {Open-vocabulary Object Detection via Vision and Language Knowledge Distillation},
  author    = {Gu, Xiuye and Lin, Tsung-Yi and Kuo, Weicheng and Cui, Yin},
  booktitle = {International Conference on Learning Representations (ICLR)},
  year      = {2022}
}

@Article{Hao2025yolov8,
    AUTHOR = {Hao, Hexiang and Peng, Yueping and Ye, Zecong and Han, Baixuan and Zhang, Xuekai and Tang, Wei and Kang, Wenchao and Li, Qilong},
    TITLE = {A High Performance Air-to-Air Unmanned Aerial Vehicle Target Detection Model},
    JOURNAL = {Drones},
    VOLUME = {9},
    YEAR = {2025},
    NUMBER = {2},
    ARTICLE-NUMBER = {154},
    URL = {https://www.mdpi.com/2504-446X/9/2/154},
    ISSN = {2504-446X},
    DOI = {10.3390/drones9020154}
}

@inproceedings{ho2020denoising,
  title     = {Denoising Diffusion Probabilistic Models},
  author    = {Ho, Jonathan and Jain, Ajay and Abbeel, Pieter},
  booktitle = {Advances in Neural Information Processing Systems (NeurIPS)},
  year      = {2020}
}

@INPROCEEDINGS{Huang2021SiamSTA,
  author={Huang, Bo and Chen, Junjie and Xu, Tingfa and Wang, Ying and Jiang, Shenwang and Wang, Yuncheng and Wang, Lei and Li, Jianan},
  booktitle={2021 IEEE/CVF International Conference on Computer Vision Workshops (ICCVW)}, 
  title={SiamSTA: Spatio-Temporal Attention based Siamese Tracker for Tracking UAVs}, 
  year={2021},
  volume={},
  number={},
  pages={1204-1212},
  doi={10.1109/ICCVW54120.2021.00140}}

@article{Huang2023SiamSRT,
  title     = {Searching Region-Free and Template-Free Siamese Network for Tracking Drones in TIR Videos},
  author    = {Bo Huang and Zeyang Dou and Junjie Chen and Jianan Li and Ning Shen and Ying Wang and Tingfa Xu},
  journal   = {IEEE Transactions on Geoscience and Remote Sensing},
  volume    = {62},
  pages     = {1--15},
  year      = {2023},
  doi       = {10.1109/TGRS.2023.3234365},
  url       = {https://ieeexplore.ieee.org/document/10012345}
}

@article{Huang2023AntiUAV410,
  title     = {Anti-UAV410: A Thermal Infrared Benchmark and Customized Scheme for Tracking Drones in the Wild},
  author    = {Bo Huang and Jianan Li and Junjie Chen and Gang Wang and Jian Zhao and Tingfa Xu},
  journal   = {IEEE Transactions on Pattern Analysis and Machine Intelligence},
  volume    = {46},
  number    = {6},
  pages     = {2852--2865},
  year      = {2024},
  doi       = {10.1109/TPAMI.2023.3237159}
}

@Article{Huang_2025_Drones,
  title   = {Expanding Open-Vocabulary Understanding for UAV Aerial Imagery: A Vision--Language Framework to Semantic Segmentation},
  author  = {Huang, Bangju and Li, Junhui and Luan, Wuyang and Tan, Jintao and Li, Chenglong and Huang, Longyang},
  journal = {Drones},
  volume  = {9},
  number  = {2},
  pages   = {155},
  year    = {2025},
  doi     = {10.3390/drones9020155}
}

@Article{huang2024EDGSYOLOv8,
    AUTHOR = {Huang, Min and Mi, Wenkai and Wang, Yuming},
    TITLE = {EDGS-YOLOv8: An Improved YOLOv8 Lightweight UAV Detection Model},
    JOURNAL = {Drones},
    VOLUME = {8},
    YEAR = {2024},
    NUMBER = {7},
    ARTICLE-NUMBER = {337},
    URL = {https://www.mdpi.com/2504-446X/8/7/337},
    ISSN = {2504-446X},
    DOI = {10.3390/drones8070337}
}

@inproceedings{Huang2023LanguageIN,
 author = {Huang, Shaohan and Dong, Li and Wang, Wenhui and Hao, Yaru and Singhal, Saksham and Ma, Shuming and Lv, Tengchao and Cui, Lei and Mohammed, Owais Khan and Patra, Barun and Liu, Qiang and Aggarwal, Kriti and Chi, Zewen and Bjorck, Nils and Chaudhary, Vishrav and Som, Subhojit  and SONG, XIA and Wei, Furu},
 booktitle = {Advances in Neural Information Processing Systems},
 editor = {A. Oh and T. Naumann and A. Globerson and K. Saenko and M. Hardt and S. Levine},
 pages = {72096--72109},
 publisher = {Curran Associates, Inc.},
 title = {Language Is Not All You Need: Aligning Perception with Language Models},
 url = {https://proceedings.neurips.cc/paper_files/paper/2023/file/e425b75bac5742a008d643826428787c-Paper-Conference.pdf},
 volume = {36},
 year = {2023}
}

@misc{Hui1,
  author       = {Bingwei Hui and Zhiyong Song and Hongqi Fan and Ping Zhong and Weidong Hu and Xiaofeng Zhang and Jianguo Lin and Hongyan Su and Wei Jin and Yongjie Zhang and Yaxi Bai},
  title        = {{A dataset for infrared image dim-small aircraft target detection and tracking under ground / air background}},
  year         = 2019,
  month        = oct,
  publisher    = {Science Data Bank},
  version      = {V1},
  doi          = {10.11922/sciencedb.902}
}

@INPROCEEDINGS{Inani2023,
  author={Inani, Kalit Naresh and Sangwan, K.S. and Dhiraj},
  booktitle={2023 4th International Conference on Innovative Trends in Information Technology (ICITIIT)}, 
  title={Machine Learning based framework for Drone Detection and Identification using RF signals}, 
  year={2023},
  volume={},
  number={},
  pages={1-8},
  doi={10.1109/ICITIIT57246.2023.10068637}
}

@ARTICLE{anti_uav,
  author={Jiang, Nan and Wang, Kuiran and Peng, Xiaoke and Yu, Xuehui and Wang, Qiang and Xing, Junliang and Li, Guorong and Guo, Guodong and Ye, Qixiang and Jiao, Jianbin and Zhao, Jian and Han, Zhenjun},
  journal={IEEE Transactions on Multimedia}, 
  title={Anti-UAV: A Large-Scale Benchmark for Vision-Based UAV Tracking}, 
  year={2023},
  volume={25},
  number={},
  pages={486-500},
  doi={10.1109/TMM.2021.3128047}}

@article{lu2023machine,
  title={Machine learning for synthetic data generation: a review},
  author={Lu, Yingzhou and Shen, Minjie and Wang, Huazheng and Wang, Xiao and van Rechem, Capucine and Fu, Tianfan and Wei, Wenqi},
  journal={arXiv preprint arXiv:2302.04062},
  year={2023}
}

@InProceedings{Kamath_2021_ICCV,
  author={Kamath, Aishwarya and Singh, Mannat and LeCun, Yann and Synnaeve, Gabriel and Misra, Ishan and Carion, Nicolas},
  booktitle={2021 IEEE/CVF International Conference on Computer Vision (ICCV)}, 
  title={MDETR - Modulated Detection for End-to-End Multi-Modal Understanding}, 
  year={2021},
  volume={},
  number={},
  pages={1760-1770},
  doi={10.1109/ICCV48922.2021.00180}
}

@INPROCEEDINGS{Katta2022,
  author={Katta, Sai Srinadhu and Nandyala, Sivaprasad and Viegas, Eduardo Kugler and AlMahmoud, Abdelrahman},
  booktitle={2022 Workshop on Benchmarking Cyber-Physical Systems and Internet of Things (CPS-IoTBench)}, 
  title={Benchmarking Audio-based Deep Learning Models for Detection and Identification of Unmanned Aerial Vehicles}, 
  year={2022},
  volume={},
  number={},
  pages={7-11},
  doi={10.1109/CPS-IoTBench56135.2022.00008}}

@article{kaufmann2023champion,
  title={Champion-level drone racing using deep reinforcement learning},
  author={Kaufmann, Elia and Bauersfeld, Leonard and Loquercio, Antonio and M{\"u}ller, Matthias and Koltun, Vladlen and Scaramuzza, Davide},
  journal={Nature},
  volume={620},
  number={7976},
  pages={982--987},
  year={2023},
  publisher={Nature Publishing Group UK London}
}

@INPROCEEDINGS{10473588,
  author={Kim, Juann and Wang, Mia Y. and Matson, Eric T.},
  booktitle={2023 Seventh IEEE International Conference on Robotic Computing (IRC)}, 
  title={Self-Supervised Drone Detection Using Acoustic Data}, 
  year={2023},
  volume={},
  number={},
  pages={67-70},
  doi={10.1109/IRC59093.2023.00018}
}

@InProceedings{Kim_2019_ICCV,
author = {Kim, Seunghyeon and Choi, Jaehoon and Kim, Taekyung and Kim, Changick},
title = {Self-Training and Adversarial Background Regularization for Unsupervised Domain Adaptive One-Stage Object Detection},
booktitle = {Proceedings of the IEEE/CVF International Conference on Computer Vision (ICCV)},
month = {October},
year = {2019}
}

@InProceedings{Kim_2019_CVPR,
  author    = {Kim, Taekyung and Jeong, Minki and Kim, Seunghyeon and Choi, Seokeon and Kim, Changick},
  title     = {Diversify and Match: A Domain Adaptive Representation Learning Paradigm for Object Detection},
  booktitle = {Proceedings of the IEEE/CVF Conference on Computer Vision and Pattern Recognition (CVPR)},
  year      = {2019},
  pages     = {12456--12465}
}

@InProceedings{Lai_2020_CVPR,
  author={Lai, Zihang and Lu, Erika and Xie, Weidi},
  booktitle={2020 IEEE/CVF Conference on Computer Vision and Pattern Recognition (CVPR)}, 
  title={MAST: A Memory-Augmented Self-Supervised Tracker}, 
  year={2020},
  volume={},
  number={},
  pages={6478-6487},
  doi={10.1109/CVPR42600.2020.00651}
}

@inproceedings{Li2023BLIP2BL,
  author = {Li, Junnan and Li, Dongxu and Savarese, Silvio and Hoi, Steven},
title = {BLIP-2: bootstrapping language-image pre-training with frozen image encoders and large language models},
year = {2023},
booktitle = {Proceedings of the 40th International Conference on Machine Learning},
articleno = {814},
numpages = {13},
location = {Honolulu, Hawaii, USA},
series = {ICML'23}
}

@InProceedings{Li_2022_CVPR,
  author={Li, Liunian Harold and Zhang, Pengchuan and Zhang, Haotian and Yang, Jianwei and Li, Chunyuan and Zhong, Yiwu and Wang, Lijuan and Yuan, Lu and Zhang, Lei and Hwang, Jenq-Neng and Chang, Kai-Wei and Gao, Jianfeng},
  booktitle={2022 IEEE/CVF Conference on Computer Vision and Pattern Recognition (CVPR)}, 
  title={Grounded Language-Image Pre-training}, 
  year={2022},
  volume={},
  number={},
  pages={10955-10965},
  doi={10.1109/CVPR52688.2022.01069}
}

@article{li2021bissiam,
  author={Li, Taotao and Hong, Zhen and Cai, Qianming and Yu, Li and Wen, Zhenyu and Yang, Renyu},
  journal={IEEE Transactions on Knowledge and Data Engineering}, 
  title={BisSiam: Bispectrum Siamese Network Based Contrastive Learning for UAV Anomaly Detection}, 
  year={2023},
  volume={35},
  number={12},
  pages={12109-12124},
  doi={10.1109/TKDE.2021.3118727}
}

@InProceedings{Li_2024_ECCV,
author="Li, Yan
and Guo, Weiwei
and Yang, Xue
and Liao, Ning
and He, Dunyun
and Zhou, Jiaqi
and Yu, Wenxian",
editor="Leonardis, Ale{\v{s}}
and Ricci, Elisa
and Roth, Stefan
and Russakovsky, Olga
and Sattler, Torsten
and Varol, G{\"u}l",
title="Toward Open Vocabulary Aerial Object Detection with CLIP-Activated Student-Teacher Learning",
booktitle="Computer Vision -- ECCV 2024",
year="2025",
publisher="Springer Nature Switzerland",
address="Cham",
pages="431--448",
isbn="978-3-031-73016-0"
}

@misc{lillicrap2015continuous,
  title={Continuous control with deep reinforcement learning}, 
      author={Timothy P. Lillicrap and Jonathan J. Hunt and Alexander Pritzel and Nicolas Heess and Tom Erez and Yuval Tassa and David Silver and Daan Wierstra},
      year={2019},
      eprint={1509.02971},
      archivePrefix={arXiv},
      primaryClass={cs.LG},
      url={https://arxiv.org/abs/1509.02971}, 
}

@inproceedings{liu2017audioCamera,
  author={Liu, Hao and Wei, Zhiqiang and Chen, Yitong and Pan, Jie and Lin, Le and Ren, Yunfang},
  booktitle={2017 IEEE Third International Conference on Multimedia Big Data (BigMM)}, 
  title={Drone Detection Based on an Audio-Assisted Camera Array}, 
  year={2017},
  volume={},
  number={},
  pages={402-406},
  doi={10.1109/BigMM.2017.57}
}

@Article{Liu_2024_ESWA,
  title = {Shooting condition insensitive unmanned aerial vehicle object detection},
journal = {Expert Systems with Applications},
volume = {246},
pages = {123221},
year = {2024},
issn = {0957-4174},
doi = {https://doi.org/10.1016/j.eswa.2024.123221},
url = {https://www.sciencedirect.com/science/article/pii/S0957417424000861},
author = {Jie Liu and Jinzong Cui and Mao Ye and Xiatian Zhu and Song Tang},
}

@ARTICLE{wang2024target,
  author={Wang, Chuanyun and Meng, Linlin and Gao, Qian and Wang, Tian and Wang, Jingjing and Wang, Linlin},
  journal={IEEE Sensors Journal}, 
  title={A Target Sensing and Visual Tracking Method for Countering Unmanned Aerial Vehicle Swarm}, 
  year={2024},
  volume={24},
  number={19},
  pages={30340-30351},
  doi={10.1109/JSEN.2024.3435856}}

@INPROCEEDINGS{10683145,
  author={Lu, Gejiacheng and Fu, Xue and Wang, Juzhen and Huang, Hao and Wang, Yu and Lin, Yun and Gui, Guan},
  booktitle={2024 IEEE 99th Vehicular Technology Conference (VTC2024-Spring)}, 
  title={A Novel Semi-Supervised Learning Method Using Self-Adaptive Threshold for UAV Recognition}, 
  year={2024},
  volume={},
  number={},
  pages={1-5},
  doi={10.1109/VTC2024-Spring62846.2024.10683145}
}

@inproceedings{Medaiyese2021,
  title={Machine learning framework for RF-based drone detection and identification system},
  author={Medaiyese, Olusiji O and Syed, Abbas and Lauf, Adrian P},
  booktitle={2021 2nd International Conference On Smart Cities, Automation \& Intelligent Computing Systems (ICON-SONICS)},
  pages={58--64},
  year={2021},
  organization={IEEE}
}

@INPROCEEDINGS{mendis2016deep,
  author={Mendis, Gihan J. and Randeny, Tharindu and Jin Wei and Madanayake, Arjuna},
  booktitle={MILCOM 2016 - 2016 IEEE Military Communications Conference}, 
  title={Deep learning based doppler radar for micro UAS detection and classification}, 
  year={2016},
  volume={},
  number={},
  pages={924-929},
  doi={10.1109/MILCOM.2016.7795448}}

@inproceedings{messina2019radar,
  title     = {Classification of Drones with a Surveillance Radar Signal},
  author    = {Messina, Marco and Pinelli, Gianpaolo},
  booktitle = {Computer Vision Systems – 12th International Conference, {ICVS} 2019, Thessaloniki, Greece, Proceedings},
  series    = {Lecture Notes in Computer Science},
  volume    = {11754},
  pages     = {723--733},
  year      = {2019},
  publisher = {Springer},
  doi       = {10.1007/978-3-030-34995-0_66}
}

@article{mnih2015human,
  title={Human-level control through deep reinforcement learning},
  author={Mnih, Volodymyr and Kavukcuoglu, Koray and Silver, David and Rusu, Andrei A and Veness, Joel and Bellemare, Marc G and Graves, Alex and Riedmiller, Martin and Fidjeland, Andreas K and Ostrovski, Georg and others},
  journal={nature},
  volume={518},
  number={7540},
  pages={529--533},
  year={2015},
  publisher={Nature Publishing Group}
}

@ARTICLE{Mou2024luster,
   author={Mou, Zhiyu and Gao, Feifei and Liu, Jun and Yun, Xiang and Wu, Qihui},
  journal={IEEE Transactions on Signal Processing}, 
  title={Cluster Head Detection for Hierarchical UAV Swarm With Graph Self-Supervised Learning}, 
  year={2024},
  volume={72},
  number={},
  pages={5517-5532},
  doi={10.1109/TSP.2024.3394654}
}

@Article{nemer2021,
    AUTHOR = {Nemer, Ibrahim and Sheltami, Tarek and Ahmad, Irfan and Yasar, Ansar Ul-Haque and Abdeen, Mohammad A. R.},
    TITLE = {RF-Based UAV Detection and Identification Using Hierarchical Learning Approach},
    JOURNAL = {Sensors},
    VOLUME = {21},
    YEAR = {2021},
    NUMBER = {6},
    ARTICLE-NUMBER = {1947},
    URL = {https://www.mdpi.com/1424-8220/21/6/1947},
    PubMedID = {33802189},
    ISSN = {1424-8220},
    DOI = {10.3390/s21061947}
}

@article{pan2010survey,
  author={Pan, Sinno Jialin and Yang, Qiang},
  journal={IEEE Transactions on Knowledge and Data Engineering}, 
  title={A Survey on Transfer Learning}, 
  year={2010},
  volume={22},
  number={10},
  pages={1345-1359},
  doi={10.1109/TKDE.2009.191}}

@inproceedings{park2015radarAudio,
  title     = {Combination of radar and audio sensors for identification of rotor-type Unmanned Aerial Vehicles ({UAV}s)},
  author    = {Park, Seongha and Shin, Sangmi and Kim, Yongho and Matson, Eric T. and Lee, Kiseon and Kolodzy, Paul J. and Slater, J. C. and Scherreik, Matthew and Sam, M. and Gallagher, J. C. and Fox, B. R. and Hopmeier, Michael},
  booktitle = {Proceedings of the 2015 IEEE Sensors},
  address   = {Busan, South Korea},
  pages     = {1--4},
  year      = {2015},
  publisher = {IEEE},
  doi       = {10.1109/ICSENS.2015.7370533}
}

@article{qiao2025fast,
author = {Qiao, Li-Hong and Wang, Rongxuan and Shu, Yucheng and Li, Baobin and Li, Weisheng and Gao, Xinbo and Cai, Zhanchuan},
year = {2025},
month = {01},
pages = {1-1},
title = {Fast Sampling of Diffusion Models for Accelerated MRI using Dual Manifold Constraints},
volume = {PP},
journal = {IEEE Transactions on Circuits and Systems for Video Technology},
doi = {10.1109/TCSVT.2024.3525015}
}

@inproceedings{Radford2021LearningTV,
  title={Learning Transferable Visual Models From Natural Language Supervision},
  author={Alec Radford and Jong Wook Kim and Chris Hallacy and Aditya Ramesh and Gabriel Goh and Sandhini Agarwal and Girish Sastry and Amanda Askell and Pamela Mishkin and Jack Clark and Gretchen Krueger and Ilya Sutskever},
  booktitle={International Conference on Machine Learning},
  year={2021},
  url={https://api.semanticscholar.org/CorpusID:231591445}
}

@misc{ramesh2022dalle2,
  title       = {Hierarchical Text-Conditional Image Generation with CLIP Latents},
  author      = {Ramesh, Aditya and Dhariwal, Prafulla and Nichol, Alexander and Chu, Casey and Chen, Mark},
  archivePrefix = {arXiv},
  eprint      = {2204.06125},
  primaryClass = {cs.CV},
  year        = {2022}
}

@inproceedings{rombach2022highresolution,
  author    = {Rombach, Robin and Blattmann, Andreas and Lorenz, Dominik and Esser, Patrick and Ommer, Bj\"orn},
    title     = {High-Resolution Image Synthesis With Latent Diffusion Models},
    booktitle = {Proceedings of the IEEE/CVF Conference on Computer Vision and Pattern Recognition (CVPR)},
    month     = {June},
    year      = {2022},
    pages     = {10684-10695}
}

@inproceedings{sadeghi2016cad2rl,
  title={{CAD2RL}: Real Single-Image Flight without a Single Real Image},
	  author={Sadeghi, Fereshteh and Levine, Sergey},
	  booktitle={Robotics: Science and Systems(RSS)},
	  year={2017}
}

@article{cetin2022,
    TITLE = {Countering a Drone in a 3D Space: Analyzing Deep Reinforcement Learning Methods},
    AUTHOR = {Çetin, Ender and Barrado, Cristina and Pastor, Enric},
    JOURNAL = {Sensors},
    VOLUME = {22},
    YEAR = {2022},
    NUMBER = {22},
    ARTICLE-NUMBER = {8863},
    URL = {https://www.mdpi.com/1424-8220/22/22/8863},
    PubMedID = {36433460},
    ISSN = {1424-8220},
    DOI = {10.3390/s22228863}
}

@Article{cetin2020,
    AUTHOR = {Çetin, Ender and Barrado, Cristina and Pastor, Enric},
    TITLE = {Counter a Drone in a Complex Neighborhood Area by Deep Reinforcement Learning},
    JOURNAL = {Sensors},
    VOLUME = {20},
    YEAR = {2020},
    NUMBER = {8},
    ARTICLE-NUMBER = {2320},
    URL = {https://www.mdpi.com/1424-8220/20/8/2320},
    PubMedID = {32325689},
    ISSN = {1424-8220},
    DOI = {10.3390/s20082320}
}

@misc{anti_uav_600,
      title={Evidential Detection and Tracking Collaboration: New Problem, Benchmark and Algorithm for Robust Anti-UAV System}, 
      author={Xue-Feng Zhu and Tianyang Xu and Jian Zhao and Jia-Wei Liu and Kai Wang and Gang Wang and Jianan Li and Qiang Wang and Lei Jin and Zheng Zhu and Junliang Xing and Xiao-Jun Wu},
      year={2023},
      eprint={2306.15767},
      archivePrefix={arXiv},
      primaryClass={cs.CV},
      url={https://arxiv.org/abs/2306.15767}, 
}

@INPROCEEDINGS{zhou2024MssamHdc,
  author={Zhou, Yue and Jiang, Yutong and Yang, Zhonglin and Li, Xingxin and Sun, Wenchen and Zhen, Han and Wang, Ying},
  booktitle={2024 3rd International Conference on Image Processing and Media Computing (ICIPMC)}, 
  title={UAV image detection based on multi-scale spatial attention mechanism with hybrid dilated convolution}, 
  year={2024},
  volume={},
  number={},
  pages={279-284},
  doi={10.1109/ICIPMC62364.2024.10586685}}

@InProceedings{Zhou_2022_ECCV,
author="Zhou, Xingyi
and Girdhar, Rohit
and Joulin, Armand
and Kr{\"a}henb{\"u}hl, Philipp
and Misra, Ishan",
editor="Avidan, Shai
and Brostow, Gabriel
and Ciss{\'e}, Moustapha
and Farinella, Giovanni Maria
and Hassner, Tal",
title="Detecting Twenty-Thousand Classes Using Image-Level Supervision",
booktitle="Computer Vision -- ECCV 2022",
year="2022",
publisher="Springer Nature Switzerland",
address="Cham",
pages="350--368",
isbn="978-3-031-20077-9"
}

@InProceedings{Zhong_2022_CVPR,
    author    = {Zhong, Yiwu and Yang, Jianwei and Zhang, Pengchuan and Li, Chunyuan and Codella, Noel and Li, Liunian Harold and Zhou, Luowei and Dai, Xiyang and Yuan, Lu and Li, Yin and Gao, Jianfeng},
    title     = {RegionCLIP: Region-Based Language-Image Pretraining},
    booktitle = {Proceedings of the IEEE/CVF Conference on Computer Vision and Pattern Recognition (CVPR)},
    month     = {June},
    year      = {2022},
    pages     = {16793-16803}
}

@ARTICLE{zheng2021air,
  author={Zheng, Ye and Chen, Zhang and Lv, Dailin and Li, Zhixing and Lan, Zhenzhong and Zhao, Shiyu},
  journal={IEEE Robotics and Automation Letters}, 
  title={Air-to-Air Visual Detection of Micro-UAVs: An Experimental Evaluation of Deep Learning}, 
  year={2021},
  volume={6},
  number={2},
  pages={1020-1027},
  doi={10.1109/LRA.2021.3056059}}

@inproceedings{li2022dual,
  author={Li, Shaogang and Gao, Jin and Li, Liang and Wang, Gang and Wang, Yizheng and Yang, Xin},
  booktitle={2022 7th International Conference on Intelligent Computing and Signal Processing (ICSP)}, 
  title={Dual-branch Approach for Tracking UAVs with the Infrared and Inverted Infrared Image}, 
  year={2022},
  volume={},
  number={},
  pages={1803-1806},
  doi={10.1109/ICSP54964.2022.9778832}
}

@ARTICLE{Wu2024EagleEye,
  author={Wu, Tongyan and Duan, Haibin and Zeng, Zhigang},
  journal={IEEE Transactions on Instrumentation and Measurement}, 
  title={Biological Eagle-Eye-Based Correlation Filter Learning for Fast UAV Tracking}, 
  year={2024},
  volume={73},
  number={},
  pages={1-12},
  doi={10.1109/TIM.2024.3436132}}

@INPROCEEDINGS{zhao2021mdrone,
  author={Zhao, Peijun and Lu, Chris Xiaoxuan and Wang, Bing and Trigoni, Niki and Markham, Andrew},
  booktitle={2021 IEEE International Conference on Robotics and Automation (ICRA)}, 
  title={3D Motion Capture of an Unmodified Drone with Single-chip Millimeter Wave Radar}, 
  year={2021},
  volume={},
  number={},
  pages={5186-5192},
  doi={10.1109/ICRA48506.2021.9561738}}

@misc{zhao20233rd,
      title={The 3rd Anti-UAV Workshop \& Challenge: Methods and Results}, 
      author={Jian Zhao and Jianan Li and Lei Jin and Jiaming Chu and Zhihao Zhang and Jun Wang and Jiangqiang Xia and Kai Wang and Yang Liu and Sadaf Gulshad and Jiaojiao Zhao and Tianyang Xu and Xuefeng Zhu and Shihan Liu and Zheng Zhu and Guibo Zhu and Zechao Li and Zheng Wang and Baigui Sun and Yandong Guo and Shin ichi Satoh and Junliang Xing and Jane Shen Shengmei},
      year={2023},
      eprint={2305.07290},
      archivePrefix={arXiv},
      primaryClass={cs.CV},
      url={https://arxiv.org/abs/2305.07290}, 
}

@ARTICLE{dut_anti_uav,
  author={Zhao, Jie and Zhang, Jingshu and Li, Dongdong and Wang, Dong},
  journal={IEEE Transactions on Intelligent Transportation Systems}, 
  title={Vision-Based Anti-UAV Detection and Tracking}, 
  year={2022},
  volume={23},
  number={12},
  pages={25323-25334},
  doi={10.1109/TITS.2022.3177627}}

@misc{zhao20212nd,
      title={The 2nd Anti-UAV Workshop \& Challenge: Methods and Results}, 
      author={Jian Zhao and Gang Wang and Jianan Li and Lei Jin and Nana Fan and Min Wang and Xiaojuan Wang and Ting Yong and Yafeng Deng and Yandong Guo and Shiming Ge and Guodong Guo},
      year={2021},
      eprint={2108.09909},
      archivePrefix={arXiv},
      primaryClass={cs.CV},
      url={https://arxiv.org/abs/2108.09909}, 
}

@article{zhang2025domain,
title = {Domain-guided conditional diffusion model for unsupervised domain adaptation},
journal = {Neural Networks},
volume = {184},
pages = {107031},
year = {2025},
issn = {0893-6080},
doi = {https://doi.org/10.1016/j.neunet.2024.107031},
url = {https://www.sciencedirect.com/science/article/pii/S0893608024009602},
author = {Yulong Zhang and Shuhao Chen and Weisen Jiang and Yu Zhang and Jiangang Lu and James T. Kwok}
}

@InProceedings{Zareian_2021_CVPR,
  title     = {Open-Vocabulary Object Detection Using Captions},
  author    = {Zareian, Alireza and Dela Rosa, Kevin and Hu, Derek Hao and Chang, Shih-Fu},
  booktitle = {Proceedings of the IEEE/CVF Conference on Computer Vision and Pattern Recognition (CVPR)},
  year      = {2021},
  pages     = {14393--14402}
}

@inproceedings{yuan2024mmaud,
  author={Yuan, Shenghai and Yang, Yizhuo and Nguyen, Thien Hoang and Nguyen, Thien-Minh and Yang, Jianfei and Liu, Fen and Li, Jianping and Wang, Han and Xie, Lihua},
  booktitle={2024 IEEE International Conference on Robotics and Automation (ICRA)}, 
  title={MMAUD: A Comprehensive Multi-Modal Anti-UAV Dataset for Modern Miniature Drone Threats}, 
  year={2024},
  volume={},
  number={},
  pages={2745-2751},
  doi={10.1109/ICRA57147.2024.10610957}
}

@inproceedings{Yu2023UnifiedTransformer,
  author={Yu, Qianjin and Ma, Yinchao and He, Jianfeng and Yang, Dawei and Zhang, Tianzhu},
  booktitle={2023 IEEE/CVF Conference on Computer Vision and Pattern Recognition Workshops (CVPRW)}, 
  title={A Unified Transformer-based Tracker for Anti-UAV Tracking}, 
  year={2023},
  volume={},
  number={},
  pages={3036-3046},
  doi={10.1109/CVPRW59228.2023.00305}
}

@ARTICLE{ying2025visible,
  author={Ying, Xinyi and Xiao, Chao and An, Wei and Li, Ruojing and He, Xu and Li, Boyang and Cao, Xu and Li, Zhaoxu and Wang, Yingqian and Hu, Mingyuan and Xu, Qingyu and Lin, Zaiping and Li, Miao and Zhou, Shilin and Sheng, Weidong and Liu, Li},
  journal={IEEE Transactions on Pattern Analysis and Machine Intelligence}, 
  title={Visible-Thermal Tiny Object Detection: A Benchmark Dataset and Baselines}, 
  year={2025},
  volume={},
  number={},
  pages={1-8},
  doi={10.1109/TPAMI.2025.3544621}}

@article{Yang2024AVFDTI,
title = {AV-FDTI: Audio-visual fusion for drone threat identification},
journal = {Journal of Automation and Intelligence},
volume = {3},
number = {3},
pages = {144-151},
year = {2024},
issn = {2949-8554},
doi = {https://doi.org/10.1016/j.jai.2024.06.002},
url = {https://www.sciencedirect.com/science/article/pii/S2949855424000285},
author = {Yizhuo Yang and Shenghai Yuan and Jianfei Yang and Thien Hoang Nguyen and Muqing Cao and Thien-Minh Nguyen and Han Wang and Lihua Xie},
}

@inproceedings{xiao2025tame,
  title={Tame: Temporal audio-based mamba for enhanced drone trajectory estimation and classification},
  author={Xiao, Zhenyuan and Hu, Huanran and Xu, Guili and He, Junwei},
  booktitle={ICASSP 2025-2025 IEEE International Conference on Acoustics, Speech and Signal Processing (ICASSP)},
  pages={1--5},
  year={2025}
}

@ARTICLE{Wu2024multi,
  author={Wu, Guangyu and Zhou, Fuhui and Kit Wong, Kai and Li, Xiang-Yang},
  journal={IEEE Journal on Selected Areas in Communications}, 
  title={A Vehicle-Mounted Radar-Vision System for Precisely Positioning Clustering UAVs}, 
  year={2024},
  volume={42},
  number={10},
  pages={2688-2703},
  doi={10.1109/JSAC.2024.3414610}}

@Article{wisniewski2024towards,
AUTHOR = {Wisniewski, Mariusz and Rana, Zeeshan A. and Petrunin, Ivan and Holt, Alan and Harman, Stephen},
TITLE = {Towards Fully Autonomous Drone Tracking by a Reinforcement Learning Agent Controlling a Pan–Tilt–Zoom Camera},
JOURNAL = {Drones},
VOLUME = {8},
YEAR = {2024},
NUMBER = {6},
ARTICLE-NUMBER = {235},
URL = {https://www.mdpi.com/2504-446X/8/6/235},
ISSN = {2504-446X},
DOI = {10.3390/drones8060235}
}

@Article{Wang_2021_IJCV,
  author   = {Wang, Ning and Zhou, Wengang and Song, Yibing and Ma, Chao and Liu, Wei and Li, Huchuan},
  title    = {Unsupervised Deep Representation Learning for Real-Time Tracking},
  journal  = {International Journal of Computer Vision},
  year     = {2021},
  volume   = {129},
  number   = {2},
  pages    = {400--418}
}

@article{zhang2024precision,
  title={Precision in pursuit: A multi-consistency joint approach for infrared anti-UAV tracking},
  author={Zhang, Junjie and Lin, Yi and Zhou, Xin and Shi, Pangrong and Zhu, Xiaoqiang and Zeng, Dan},
  journal={The Visual Computer},
  volume = {41},
  pages={2187–2202},
  year={2024},
  publisher={Springer}
}

@Article{wang2022uavswarm,
AUTHOR = {Wang, Chuanyun and Su, Yang and Wang, Jingjing and Wang, Tian and Gao, Qian},
TITLE = {UAVSwarm Dataset: An Unmanned Aerial Vehicle Swarm Dataset for Multiple Object Tracking},
JOURNAL = {Remote Sensing},
VOLUME = {14},
YEAR = {2022},
NUMBER = {11},
ARTICLE-NUMBER = {2601},
URL = {https://www.mdpi.com/2072-4292/14/11/2601},
ISSN = {2072-4292},
DOI = {10.3390/rs14112601}
}

@ARTICLE{tu2024,
  author={Tu, Xiaohan and Zhang, Chuanhao and Zhuang, Haiyan and Liu, Siping and Li, Renfa},
  journal={IEEE Access}, 
  title={Fast Drone Detection With Optimized Feature Capture and Modeling Algorithms}, 
  year={2024},
  volume={12},
  number={},
  pages={108374-108388},
  doi={10.1109/ACCESS.2024.3438991}
}

@InProceedings{tsai2018learning,
author = {Tsai, Yi-Hsuan and Hung, Wei-Chih and Schulter, Samuel and Sohn, Kihyuk and Yang, Ming-Hsuan and Chandraker, Manmohan},
title = {Learning to Adapt Structured Output Space for Semantic Segmentation},
booktitle = {Proceedings of the IEEE Conference on Computer Vision and Pattern Recognition (CVPR)},
month = {June},
year = {2018}
}

@INPROCEEDINGS{torralba2011unbiased,
  author={Torralba, Antonio and Efros, Alexei A.},
  booktitle={CVPR 2011}, 
  title={Unbiased look at dataset bias}, 
  year={2011},
  volume={},
  number={},
  pages={1521-1528},
  doi={10.1109/CVPR.2011.5995347}}

@inproceedings{halmstad_drone,
  author={Svanström, Fredrik and Englund, Cristofer and Alonso-Fernandez, Fernando},
  booktitle={2020 25th International Conference on Pattern Recognition (ICPR)}, 
  title={Real-Time Drone Detection and Tracking With Visible, Thermal and Acoustic Sensors}, 
  year={2021},
  volume={},
  number={},
  pages={7265-7272},
  doi={10.1109/ICPR48806.2021.9413241}
}

@article{Sun2024multiyolov8,
    author = {Sun, Shizun and Mo, Bo and Xu, Junwei and Li, Dawei and Zhao, Jie and Han, Shuo},
title = {Multi-YOLOv8: An infrared moving small object detection model based on YOLOv8 for air vehicle},
year = {2024},
issue_date = {Jul 2024},
publisher = {Elsevier Science Publishers B. V.},
address = {NLD},
volume = {588},
number = {C},
issn = {0925-2312},
url = {https://doi.org/10.1016/j.neucom.2024.127685},
doi = {10.1016/j.neucom.2024.127685},
journal = {Neurocomput.},
month = jul,
numpages = {24}
}

@inproceedings{Su2019VLBERTPO,
  title={VL-BERT: Pre-training of Generic Visual-Linguistic Representations},
  author={Weijie Su and Xizhou Zhu and Yue Cao and Bin Li and Lewei Lu and Furu Wei and Jifeng Dai},
  booktitle = {International Conference on Learning Representations (ICLR)},
  year={2020}
}

@inproceedings{song2021score,
  title     = {Score-Based Generative Modeling through Stochastic Differential Equations},
  author    = {Song, Yang and Ermon, Stefano},
  booktitle = {International Conference on Learning Representations (ICLR)},
  year      = {2021}
}

@inproceedings{sohl2015deep,
  author = {Sohl-Dickstein, Jascha and Weiss, Eric A. and Maheswaranathan, Niru and Ganguli, Surya},
title = {Deep unsupervised learning using nonequilibrium thermodynamics},
year = {2015},
booktitle = {Proceedings of the 32nd International Conference on International Conference on Machine Learning - Volume 37},
pages = {2256–2265},
numpages = {10},
location = {Lille, France},
series = {ICML'15}
}

@InProceedings{Sio_2020_MM,
  author = {Sio, Chon Hou and Ma, Yu-Jen and Shuai, Hong-Han and Chen, Jun-Cheng and Cheng, Wen-Huang},
title = {S2SiamFC: Self-supervised Fully Convolutional Siamese Network for Visual Tracking},
year = {2020},
isbn = {9781450379885},
publisher = {Association for Computing Machinery},
address = {New York, NY, USA},
url = {https://doi.org/10.1145/3394171.3413611},
doi = {10.1145/3394171.3413611},
booktitle = {Proceedings of the 28th ACM International Conference on Multimedia},
pages = {1948–1957},
numpages = {10},
location = {Seattle, WA, USA},
series = {MM '20}
}

@article{silver2016mastering,
  title={Mastering the game of Go with deep neural networks and tree search},
  author={Silver, David and Huang, Aja and Maddison, Chris J and Guez, Arthur and Sifre, Laurent and Van Den Driessche, George and Schrittwieser, Julian and Antonoglou, Ioannis and Panneershelvam, Veda and Lanctot, Marc and others},
  journal={Nature},
  volume={529},
  number={7587},
  pages={484--489},
  year={2016},
  publisher={Nature Publishing Group}
}

@article{samaras2019multisensor,
AUTHOR = {Samaras, Stamatios and Diamantidou, Eleni and Ataloglou, Dimitrios and Sakellariou, Nikos and Vafeiadis, Anastasios and Magoulianitis, Vasilis and Lalas, Antonios and Dimou, Anastasios and Zarpalas, Dimitrios and Votis, Konstantinos and Daras, Petros and Tzovaras, Dimitrios},
TITLE = {Deep Learning on Multi Sensor Data for Counter UAV Applications—A Systematic Review},
JOURNAL = {Sensors},
VOLUME = {19},
YEAR = {2019},
NUMBER = {22},
ARTICLE-NUMBER = {4837},
URL = {https://www.mdpi.com/1424-8220/19/22/4837},
PubMedID = {31698862},
ISSN = {1424-8220},
DOI = {10.3390/s19224837}
}

@Article{wang2024survey,
AUTHOR = {Wang, Bingshu and Li, Qiang and Mao, Qianchen and Wang, Jinbao and Chen, C. L. Philip and Shangguan, Aihong and Zhang, Haosu},
TITLE = {A Survey on Vision-Based Anti Unmanned Aerial Vehicles Methods},
JOURNAL = {Drones},
VOLUME = {8},
YEAR = {2024},
NUMBER = {9},
ARTICLE-NUMBER = {518},
URL = {https://www.mdpi.com/2504-446X/8/9/518},
ISSN = {2504-446X},
DOI = {10.3390/drones8090518}
}

@article{svanstrom2021dataset,
  title={A dataset for multi-sensor drone detection},
  author={Svanstr{\"o}m, Fredrik and Alonso-Fernandez, Fernando and Englund, Cristofer},
  journal={Data in Brief},
  volume={39},
  pages={107521},
  year={2021},
  publisher={Elsevier}
}

@inproceedings{benigmim2023one,
  title={One-shot unsupervised domain adaptation with personalized diffusion models},
  author={Benigmim, Yasser and Roy, Subhankar and Essid, Slim and Kalogeiton, Vicky and Lathuili{\`e}re, St{\'e}phane},
  booktitle={Proceedings of the IEEE/CVF conference on computer vision and pattern recognition},
  pages={698--708},
  year={2023}
}

@InProceedings{Ye_2022_CVPR,
  author    = {Ye, Junjie and Fu, Changhong and Zheng, Guangze and Paudel, Danda Pani and Chen, Guang},
  title     = {Unsupervised Domain Adaptation for Nighttime Aerial Tracking},
  booktitle = {Proceedings of the IEEE/CVF Conference on Computer Vision and Pattern Recognition (CVPR)},
  year      = {2022},
  pages     = {8896--8905}
}

@InProceedings{Feng_2024_NeurIPS,
  title     = {MemVLT: Vision-Language Tracking with Adaptive Memory-based Prompts},
  author    = {Feng, Xiaokun and Li, Xuchen and Hu, Shiyu and Zhang, Dailing and Wu, Meiqi and Zhang, Jing and Chen, Xiaotang and Huang, Kaiqi},
  booktitle = {Advances in Neural Information Processing Systems 37 (NeurIPS 2024)},
  year      = {2024}
}

@inproceedings{salimans2022progressive,
  title     = {Progressive Distillation for Fast Sampling of Diffusion Models},
  author    = {Salimans, Tim and Ho, Jonathan},
  booktitle = {International Conference on Learning Representations (ICLR)},
  year      = {2022}
}

@inproceedings{saharia2022imagen,
  author = {Saharia, Chitwan and Chan, William and Saxena, Saurabh and Lit, Lala and Whang, Jay and Denton, Emily and Ghasemipour, Seyed Kamyar Seyed and Ayan, Burcu Karagol and Mahdavi, S. Sara and Gontijo-Lopes, Raphael and Salimans, Tim and Ho, Jonathan and Fleet, David J and Norouzi, Mohammad},
title = {Photorealistic text-to-image diffusion models with deep language understanding},
year = {2022},
isbn = {9781713871088},
publisher = {Curran Associates Inc.},
address = {Red Hook, NY, USA},
booktitle = {Proceedings of the 36th International Conference on Neural Information Processing Systems},
articleno = {2643},
numpages = {16},
location = {New Orleans, LA, USA},
series = {NIPS '22}
}

@Article{Wisniewski2022,
AUTHOR = {Wisniewski, Mariusz and Rana, Zeeshan A. and Petrunin, Ivan},
TITLE = {Drone Model Classification Using Convolutional Neural Network Trained on Synthetic Data},
JOURNAL = {Journal of Imaging},
VOLUME = {8},
YEAR = {2022},
NUMBER = {8},
ARTICLE-NUMBER = {218},
URL = {https://www.mdpi.com/2313-433X/8/8/218},
PubMedID = {36005461},
ISSN = {2313-433X},
DOI = {10.3390/jimaging8080218}
}

@dataset{Wisniewski2022dataset,
  author       = {Mariusz Wisniewski and Zeeshan A. Rana and Ivan Petrunin},
  title        = {Synthetic Drone Classification Dataset},
  year         = {2022},
  month        = {Jul},
  day          = {28},
  publisher    = {Cranfield University},
  doi          = {10.17862/cranfield.rd.19423925},
  url          = {https://doi.org/10.17862/cranfield.rd.19423925},
  license      = {CC BY 4.0},
  rights       = {Creative Commons Attribution 4.0 International},
}

@ARTICLE{Scholes2022,
  author={Scholes, Stirling and Ruget, Alice and Mora-Martín, Germán and Zhu, Feng and Gyongy, Istvan and Leach, Jonathan},
  journal={IEEE Access}, 
  title={DroneSense: The Identification, Segmentation, and Orientation Detection of Drones via Neural Networks}, 
  year={2022},
  volume={10},
  number={},
  pages={38154-38164},
  doi={10.1109/ACCESS.2022.3162866}
}

@ARTICLE{Wu2021Survey,
  author={Wu, Xin and Li, Wei and Hong, Danfeng and Tao, Ran and Du, Qian},
  journal={IEEE Geoscience and Remote Sensing Magazine}, 
  title={Deep Learning for Unmanned Aerial Vehicle-Based Object Detection and Tracking: A survey}, 
  year={2022},
  volume={10},
  number={1},
  pages={91-124},
  doi={10.1109/MGRS.2021.3115137}}

@inproceedings{chen2021transformer,
  title={Transformer tracking},
  author={Chen, Xin and Yan, Bin and Zhu, Jiawen and Wang, Dong and Yang, Xiaoyun and Lu, Huchuan},
  booktitle={Proceedings of the IEEE/CVF conference on computer vision and pattern recognition},
  pages={8126--8135},
  year={2021}
}

@inproceedings{yan2021learning,
  author    = {Yan, Bin and Peng, Houwen and Fu, Jianlong and Wang, Dong and Lu, Huchuan},
    title     = {Learning Spatio-Temporal Transformer for Visual Tracking},
    booktitle = {Proceedings of the IEEE/CVF International Conference on Computer Vision (ICCV)},
    month     = {October},
    year      = {2021},
    pages     = {10448-10457}
}

@ARTICLE{xiao2024target,
AUTHOR={Xiao, Dingkun  and Wei, Zhenzhong  and Zhang, Guangjun },
TITLE={Target-aware transformer tracking with hard occlusion instance generation},
JOURNAL={Frontiers in Neurorobotics},
VOLUME={17},
YEAR={2024},
URL={https://www.frontiersin.org/journals/neurorobotics/articles/10.3389/fnbot.2023.1323188},
DOI={10.3389/fnbot.2023.1323188},
ISSN={1662-5218}}

@misc{xiao2024av,
      title={AV-DTEC: Self-Supervised Audio-Visual Fusion for Drone Trajectory Estimation and Classification}, 
      author={Zhenyuan Xiao and Yizhuo Yang and Guili Xu and Xianglong Zeng and Shenghai Yuan},
      year={2024},
      eprint={2412.16928},
      archivePrefix={arXiv},
      primaryClass={cs.SD},
      url={https://arxiv.org/abs/2412.16928}, 
}

@article{shen2022real,
  author = {Shen, Hao and Lin, Defu and Song, Tao},
title = {A real-time siamese tracker deployed on UAVs},
year = {2022},
issue_date = {Apr 2022},
publisher = {Springer-Verlag},
address = {Berlin, Heidelberg},
volume = {19},
number = {2},
issn = {1861-8200},
url = {https://doi.org/10.1007/s11554-021-01190-z},
doi = {10.1007/s11554-021-01190-z},
journal = {J. Real-Time Image Process.},
month = apr,
pages = {463–473},
numpages = {11}
}

@article{do2025ramots,
  title={RAMOTS: A Real-Time System for Aerial Multi-Object Tracking based on Deep Learning and Big Data Technology},
  author={Do, Nhat-Tan and Nguyen, Nhi Ngoc-Yen and Nguyen, Dieu-Phuong and Do, Trong-Hop},
  journal={arXiv preprint arXiv:2502.03760},
  year={2025}
}

@inproceedings{li2019multi,
  author={Li, Jing and Ye, Dong Hye and Chung, Timothy and Kolsch, Mathias and Wachs, Juan and Bouman, Charles},
  booktitle={2016 IEEE/RSJ International Conference on Intelligent Robots and Systems (IROS)}, 
  title={Multi-target detection and tracking from a single camera in Unmanned Aerial Vehicles (UAVs)}, 
  year={2016},
  volume={},
  number={},
  pages={4992-4997},
  doi={10.1109/IROS.2016.7759733}
}

@misc{tang2024revisiting,
      title={Revisiting RGBT Tracking Benchmarks from the Perspective of Modality Validity: A New Benchmark, Problem, and Solution}, 
      author={Zhangyong Tang and Tianyang Xu and Zhenhua Feng and Xuefeng Zhu and Chunyang Cheng and Xiao-Jun Wu and Josef Kittler},
      year={2025},
      eprint={2405.00168},
      archivePrefix={arXiv},
      primaryClass={cs.CV},
      url={https://arxiv.org/abs/2405.00168}, 
}

@inproceedings{cui2022mixformer,
  title={Mixformer: End-to-end tracking with iterative mixed attention},
  author={Cui, Yutao and Jiang, Cheng and Wang, Limin and Wu, Gangshan},
  booktitle={Proceedings of the IEEE/CVF conference on computer vision and pattern recognition},
  pages={13608--13618},
  year={2022}
}

@InProceedings{zhang2022bytetrack,
author="Zhang, Yifu
and Sun, Peize
and Jiang, Yi
and Yu, Dongdong
and Weng, Fucheng
and Yuan, Zehuan
and Luo, Ping
and Liu, Wenyu
and Wang, Xinggang",
editor="Avidan, Shai
and Brostow, Gabriel
and Ciss{\'e}, Moustapha
and Farinella, Giovanni Maria
and Hassner, Tal",
title="ByteTrack: Multi-object Tracking by Associating Every Detection Box",
booktitle="Computer Vision -- ECCV 2022",
year="2022",
publisher="Springer Nature Switzerland",
address="Cham",
pages="1--21",
isbn="978-3-031-20047-2"
}

@ARTICLE{Elleuch2024jamming,
  author={Elleuch, Ibrahim and Pourranjbar, Ali and Kaddoum, Georges},
  journal={IEEE Open Journal of the Communications Society}, 
  title={Leveraging Transformer Models for Anti-Jamming in Heavily Attacked UAV Environments}, 
  year={2024},
  volume={5},
  number={},
  pages={5337-5347},
  doi={10.1109/OJCOMS.2024.3451288}
}

@Article{uddin2022tvddt,
AUTHOR = {Uddin, Zahoor and Qamar, Aamir and Alharbi, Abdullah G. and Orakzai, Farooq Alam and Ahmad, Ayaz},
TITLE = {Detection of Multiple Drones in a Time-Varying Scenario Using Acoustic Signals},
JOURNAL = {Sustainability},
VOLUME = {14},
YEAR = {2022},
NUMBER = {7},
ARTICLE-NUMBER = {4041},
URL = {https://www.mdpi.com/2071-1050/14/7/4041},
ISSN = {2071-1050},
DOI = {10.3390/su14074041}
}

@inproceedings{ren2016faster,
  title={Faster r-cnn: Towards real-time object detection with region proposal networks},
  author={Ren, Shaoqing and He, Kaiming and Girshick, Ross and Sun, Jian},
  booktitle={Advances in neural information processing systems},
  volume={28},
  year={2015}
}

@article{bochkovskiy2020yolov4,
  title={Yolov4: Optimal speed and accuracy of object detection},
  author={Bochkovskiy, Alexey and Wang, Chien-Yao and Liao, Hong-Yuan Mark},
  journal={arXiv preprint arXiv:2004.10934},
  year={2020}
}

@InProceedings{wang2022yolov7,
    author    = {Wang, Chien-Yao and Bochkovskiy, Alexey and Liao, Hong-Yuan Mark},
    title     = {YOLOv7: Trainable Bag-of-Freebies Sets New State-of-the-Art for Real-Time Object Detectors},
    booktitle = {Proceedings of the IEEE/CVF Conference on Computer Vision and Pattern Recognition (CVPR)},
    month     = {June},
    year      = {2023},
    pages     = {7464-7475}
}

@article{He2024FogUAV,
  author = {He, Xin and Fan, Kuangang and Xu, Zhitao},
  title = {Uav identification based on improved YOLOv7 under foggy condition},
  journal = {Signal, Image and Video Processing},
  volume = {18},
  number = {8},
  pages = {6173--6183},
  year = {2024},
  month = {September},
  day = {1},
  issn = {1863-1711},
  doi = {10.1007/s11760-024-03305-y},
  url = {https://doi.org/10.1007/s11760-024-03305-y}
}

@INPROCEEDINGS{singh2024VisionUAV,
  author={Singh, Pranita and Gupta, Keshav and Jain, Amit Kumar and Vishakha and Jain, Abhishek and Jain, Arpit},
  booktitle={2024 2nd International Conference on Disruptive Technologies (ICDT)}, 
  title={Vision-based UAV Detection in Complex Backgrounds and Rainy Conditions}, 
  year={2024},
  volume={},
  number={},
  pages={1097-1102},
  doi={10.1109/ICDT61202.2024.10489147}
}

@inproceedings{carion2020end,
  title={End-to-end object detection with transformers},
  author={Carion, Nicolas and Massa, Francisco and Synnaeve, Gabriel and Usunier, Nicolas and Kirillov, Alexander and Zagoruyko, Sergey},
  booktitle={European conference on computer vision},
  pages={213--229},
  year={2020},
  organization={Springer}
}

@inproceedings{liu2021swin,
  author={Liu, Ze and Lin, Yutong and Cao, Yue and Hu, Han and Wei, Yixuan and Zhang, Zheng and Lin, Stephen and Guo, Baining},
  booktitle={2021 IEEE/CVF International Conference on Computer Vision (ICCV)}, 
  title={Swin Transformer: Hierarchical Vision Transformer using Shifted Windows}, 
  year={2021},
  volume={},
  number={},
  pages={9992-10002},
  doi={10.1109/ICCV48922.2021.00986}}

@article{arsenos2024,
title = {NEFELI: A deep-learning detection and tracking pipeline for enhancing autonomy in advanced air mobility},
journal = {Aerospace Science and Technology},
volume = {155},
pages = {109613},
year = {2024},
issn = {1270-9638},
doi = {https://doi.org/10.1016/j.ast.2024.109613},
url = {https://www.sciencedirect.com/science/article/pii/S1270963824007429},
author = {Anastasios Arsenos and Evangelos Petrongonas and Orfeas Filippopoulos and Christos Skliros and Dimitrios Kollias and Stefanos Kollias}
}

@article{Filkin2021UnmannedAV,
  title={Unmanned Aerial Vehicles for Operational Monitoring of Landfills},
  author={Timofey Filkin and Natalia Sliusar and Marco Ritzkowski and Marion Huber-Humer},
  journal={Drones},
  year={2021},
  url={https://api.semanticscholar.org/CorpusID:240078215}
}

@misc{aot2023,
  title={Airborne Object Tracking Dataset},
  author={{Amazon}},
  howpublished={Registry of Open Data on AWS},
  note={Available at: https://registry.opendata.aws/airborne-object-tracking},
  year={2023}
}

@INPROCEEDINGS{Sangam2023TransVisDrone,
  author={Sangam, Tushar and Dave, Ishan Rajendrakumar and Sultani, Waqas and Shah, Mubarak},
  booktitle={2023 IEEE International Conference on Robotics and Automation (ICRA)}, 
  title={TransVisDrone: Spatio-Temporal Transformer for Vision-based Drone-to-Drone Detection in Aerial Videos}, 
  year={2023},
  volume={},
  number={},
  pages={6006-6013},
  doi={10.1109/ICRA48891.2023.10161433}
}

@Article{Seidaliyeva2020,
    AUTHOR = {Seidaliyeva, Ulzhalgas and Akhmetov, Daryn and Ilipbayeva, Lyazzat and Matson, Eric T.},
TITLE = {Real-Time and Accurate Drone Detection in a Video with a Static Background},
JOURNAL = {Sensors},
VOLUME = {20},
YEAR = {2020},
NUMBER = {14},
ARTICLE-NUMBER = {3856},
URL = {https://www.mdpi.com/1424-8220/20/14/3856},
PubMedID = {32664365},
ISSN = {1424-8220},
}

@misc{yoshihashi2018,
      title={Differentiating Objects by Motion: Joint Detection and Tracking of Small Flying Objects}, 
      author={Ryota Yoshihashi and Tu Tuan Trinh and Rei Kawakami and Shaodi You and Makoto Iida and Takeshi Naemura},
      year={2018},
      eprint={1709.04666},
      archivePrefix={arXiv},
      primaryClass={cs.CV},
      url={https://arxiv.org/abs/1709.04666}, 
}

@ARTICLE{mavvid2020,
  author={Rodriguez-Ramos, Alejandro and Rodriguez-Vazquez, Javier and Sampedro, Carlos and Campoy, Pascual},
  journal={IEEE Access}, 
  title={Adaptive Inattentional Framework for Video Object Detection With Reward-Conditional Training}, 
  year={2020},
  volume={8},
  number={},
  pages={124451-124466},
  doi={10.1109/ACCESS.2020.3006191}}

@misc{zhu2020vision,
      title={Vision Meets Drones: A Challenge}, 
      author={Pengfei Zhu and Longyin Wen and Xiao Bian and Haibin Ling and Qinghua Hu},
      year={2018},
      eprint={1804.07437},
      archivePrefix={arXiv},
      primaryClass={cs.CV},
      url={https://arxiv.org/abs/1804.07437}, 
}

@InProceedings{tan2020efficientdet,
author = {Tan, Mingxing and Pang, Ruoming and Le, Quoc V.},
title = {EfficientDet: Scalable and Efficient Object Detection},
booktitle = {Proceedings of the IEEE/CVF Conference on Computer Vision and Pattern Recognition (CVPR)},
month = {June},
year = {2020}
}

@InProceedings{Tan2019EfficientNetRM,
  title = 	 {{E}fficient{N}et: Rethinking Model Scaling for Convolutional Neural Networks},
  author =       {Tan, Mingxing and Le, Quoc},
  booktitle = 	 {Proceedings of the 36th International Conference on Machine Learning},
  pages = 	 {6105--6114},
  year = 	 {2019},
  editor = 	 {Chaudhuri, Kamalika and Salakhutdinov, Ruslan},
  volume = 	 {97},
  series = 	 {Proceedings of Machine Learning Research},
  month = 	 {09--15 Jun},
  url = 	 {https://proceedings.mlr.press/v97/tan19a.html}
}

@inproceedings{du2018unmanned,
  title={The unmanned aerial vehicle benchmark: Object detection and tracking},
  author={Du, Dawei and Qi, Yuankai and Yu, Hongyang and Yang, Yifan and Duan, Kaiwen and Li, Guorong and Zhang, Weigang and Huang, Qingming and Tian, Qi},
  booktitle={Proceedings of the European conference on computer vision (ECCV)},
  pages={370--386},
  year={2018}
}

@article{Paweczyk2020RealWO,
  author={Pawełczyk, Maciej Ł. and Wojtyra, Marek},
  journal={IEEE Access}, 
  title={Real World Object Detection Dataset for Quadcopter Unmanned Aerial Vehicle Detection}, 
  year={2020},
  volume={8},
  number={},
  pages={174394-174409},
  doi={10.1109/ACCESS.2020.3026192}}

@article{Fan2021DesignAI,
  title={Design and Implementation of Intelligent Inspection and Alarm Flight System for Epidemic Prevention},
  author={Jiwei Fan and Xiaogang Yang and Ruitao Lu and Xueli Xie and Weipeng Li},
  journal={Drones},
  year={2021},
  url={https://api.semanticscholar.org/CorpusID:237708299}
}

@misc{USCDroneDatasetWeb,
  author       = {University of Southern California Media Communications Lab (MCL)},
  title        = {USC Drone Detection and Tracking Dataset Webpage},
  howpublished = {\url{http://mcl.usc.edu/mcl-drone-dataset/}},
  year         = {2019},
  note         = {Accessed: Mar. 18, 2025}
}

@article{Wang2019USCDrone, 
title={Towards Visible and Thermal Drone Monitoring with Convolutional Neural Networks}, 
volume={8}, 
DOI={10.1017/ATSIP.2018.30}, 
journal={APSIPA Transactions on Signal and Information Processing}, 
author={Wang, Ye and Chen, Yueru and Choi, Jongmoo and Kuo, C.-C. Jay}, 
year={2019}, 
pages={e5}
}

@misc{HalmstadDatasetRepo,
  author       = {Fredrik Svanstr{\"o}m and Cristofer Englund and Fernando Alonso-Fernandez},
  title        = {Halmstad Drone Dataset (Multi-sensor Drone Detection Dataset) Repository},
  howpublished = {\url{https://github.com/DroneDetectionThesis/Drone-detection-dataset}},
  year         = {2021},
  note         = {GitHub repository, Accessed: Mar. 18, 2025}
}

@article{rosner2025multimodal,
  title={Multimodal dataset for indoor 3D drone tracking},
  author={Rosner, Jakub and Krzeszowski, Tomasz and {\'S}wito{\'n}ski, Adam and Josi{\'n}ski, Henryk and Lindenheim-Locher, Wojciech and Zieli{\'n}ski, Micha{\l} and Paleta, Grzegorz and Paszkuta, Marcin and Wojciechowski, Konrad},
  journal={Scientific Data},
  volume={12},
  number={1},
  pages={257},
  year={2025},
  publisher={Nature Publishing Group UK London}
}

@INPROCEEDINGS{viktor2020midgard,
  author={Walter, Viktor and Vrba, Matouš and Saska, Martin},
  booktitle={2020 IEEE International Conference on Robotics and Automation (ICRA)}, 
  title={On training datasets for machine learning-based visual relative localization of micro-scale UAVs}, 
  year={2020},
  volume={},
  number={},
  pages={10674-10680},
  doi={10.1109/ICRA40945.2020.9196947}
}

@article{andle2022stanford,
  title={The stanford drone dataset is more complex than we think: An analysis of key characteristics},
  author={Andle, Josh and Soucy, Nicholas and Socolow, Simon and Sekeh, Salimeh Yasaei},
  journal={IEEE Transactions on Intelligent Vehicles},
  volume={8},
  number={2},
  pages={1863--1873},
  year={2022},
  publisher={IEEE}
}

@inproceedings{kalra2019dronesurf,
  title={Dronesurf: Benchmark dataset for drone-based face recognition},
  author={Kalra, Isha and Singh, Maneet and Nagpal, Shruti and Singh, Richa and Vatsa, Mayank and Sujit, PB},
  booktitle={2019 14th IEEE International Conference on Automatic Face \& Gesture Recognition (FG 2019)},
  pages={1--7},
  year={2019},
  organization={IEEE}
}

@misc{AntiUAVChallenge2023,
    author       = {Anti-UAV Challenge Organizers},
    title        = {The 4th Anti-UAV Workshop and Challenge (CVPR 2025)},
    howpublished = {\url{https://anti-uav.github.io/}},
    note         = {Accessed: Mar. 18, 2025},
    year={2023},
}

@misc{UG2Challenge2024,
    author       = {UG2+ Challenge Organizers},
    title        = {The 7th UG2+ Challenge (CVPR 2024)},
    howpublished = {\url{https://ug2-uav-tracking.github.io/index.html}},
    note         = {Accessed: Mar. 22, 2025},
    year={2024},
}

@inproceedings{Nguyen2017,
author = {Nguyen, Phuc and Truong, Hoang and Ravindranathan, Mahesh and Nguyen, Anh and Han, Richard and Vu, Tam},
title = {Matthan: Drone Presence Detection by Identifying Physical Signatures in the Drone's RF Communication},
year = {2017},
isbn = {9781450349284},
publisher = {Association for Computing Machinery},
address = {New York, NY, USA},
url = {https://doi.org/10.1145/3081333.3081354},
doi = {10.1145/3081333.3081354},
booktitle = {Proceedings of the 15th Annual International Conference on Mobile Systems, Applications, and Services},
pages = {211–224},
numpages = {14},
location = {Niagara Falls, New York, USA},
series = {MobiSys '17}
}

@INPROCEEDINGS{Jeon2017,
  author={Jeon, Sungho and Shin, Jong-Woo and Lee, Young-Jun and Kim, Woong-Hee and Kwon, YoungHyoun and Yang, Hae-Yong},
  booktitle={2017 25th European Signal Processing Conference (EUSIPCO)}, 
  title={Empirical study of drone sound detection in real-life environment with deep neural networks}, 
  year={2017},
  volume={},
  number={},
  pages={1858-1862},
  doi={10.23919/EUSIPCO.2017.8081531}}

@article{Sun2023DeepLD,
  author = {Sun, Yumeng and Li, Jinguang and Wang, Linwei and Xv, Junjie and Liu, yu},
year = {2023},
month = {10},
pages = {},
title = {Deep Learning-based drone acoustic event detection system for microphone arrays},
volume = {83},
journal = {Multimedia Tools and Applications},
doi = {10.1007/s11042-023-17477-1}
}

@article{dewangan2023application,
  title={Application of image processing techniques for uav detection using deep learning and distance-wise analysis},
  author={Dewangan, Vedanshu and Saxena, Aditya and Thakur, Rahul and Tripathi, Shrivishal},
  journal={Drones},
  volume={7},
  number={3},
  pages={174},
  year={2023},
  publisher={MDPI}
}

@article{Unlu2019,
  title = {Deep learning-based strategies for the detection and tracking of drones using several cameras},
	volume = {11},
	issn = {1882-6695},
	url = {https://doi.org/10.1186/s41074-019-0059-x},
	doi = {10.1186/s41074-019-0059-x},
	number = {7},
	journal = {IPSJ Transactions on Computer Vision and Applications},
	author = {Unlu, Eren and Zenou, Emmanuel and Riviere, Nicolas and Dupouy, Paul-Edouard},
	month = jul,
	year = {2019},
}

@inproceedings{Nair2024ThermalFL,
author = {Nair, Akarsh K and Sahoo, Jayakrushna and Raj, Ebin Deni},
title = {A lightweight FL-based UAV detection model using thermal images},
year = {2024},
isbn = {9798400709296},
publisher = {Association for Computing Machinery},
address = {New York, NY, USA},
url = {https://doi.org/10.1145/3659677.3659705},
doi = {10.1145/3659677.3659705},
booktitle = {Proceedings of the 7th International Conference on Networking, Intelligent Systems and Security},
articleno = {15},
numpages = {5},
location = {Meknes, AA, Morocco},
series = {NISS '24}
}

@inproceedings{cabrera2019detection,
  title={Detection of nearby UAVs using CNN and Spectrograms},
  author={Cabrera-Ponce, Aldrich A and Martinez-Carranza, J and Rascon, Caleb},
  booktitle={Proceedings of the International Micro Air Vehicle Conference and Competition (IMAV)(Madrid), Madrid, Spain},
  volume={29},
  year={2019}
}

@article{Elsayed2024LERFNetAE,
  title={LERFNet: an enlarged effective receptive field backbone network for enhancing visual drone detection},
  author = {Elsayed, Mohamed and Reda, Mohamed and Mashaly, Ahmed S. and Amein, Ahmed S.},
  journal={Vis. Comput.},
  year={2024},
  volume={41},
  pages={2219-2232},
  url={https://api.semanticscholar.org/CorpusID:270955625}
}

@misc{wisniewski2024fastrcnn,
      title={Drone Detection using Deep Neural Networks Trained on Pure Synthetic Data}, 
      author={Mariusz Wisniewski and Zeeshan A. Rana and Ivan Petrunin and Alan Holt and Stephen Harman},
      year={2024},
      eprint={2411.09077},
      archivePrefix={arXiv},
      primaryClass={cs.CV},
      url={https://arxiv.org/abs/2411.09077}, 
}
